\documentclass[lettersize,journal]{IEEEtran}
\usepackage{amssymb}
\usepackage{amsmath,amsfonts}
\usepackage{algorithmic}
\usepackage{algorithm}
\usepackage{array}
\usepackage[caption=false,font=normalsize,labelfont=sf,textfont=sf]{subfig}
\usepackage{textcomp}
\usepackage{stfloats}
\usepackage{url}
\usepackage{verbatim}
\usepackage{graphicx}
\usepackage{multirow}
\usepackage{hyperref}
\usepackage{mathrsfs}

\hypersetup{hypertex=true,
	colorlinks=true,
	linkcolor=red,
	anchorcolor=green,
	citecolor=green}
\usepackage{cite}
\begin{document}

\title{

Learning a Single Convolutional Layer Model for  Low Light Image Enhancement

 }

\author{Yuantong Zhang, Baoxin Teng, Daiqin Yang, Zhenzhong Chen,~\IEEEmembership{Senior~Member,~IEEE} \\
  Haichuan Ma, Gang Li, 
	and Wenpeng Ding		
	\thanks{This work is supported in part by Media Innovation Lab, Architecture and Technology Innovation Department, Cloud BU, Huawei.  Y. Zhang, B. Teng, Z. Chen and D. Yang are with the school of Remote Sensing and Information Engineering, Wuhan University, Hubei 430079, China. H. Ma, G. Li, and W. Ding are with Media Innovation Lab, Architecture and Technology Innovation Department, Cloud BU, Huawei. (Corresponding author: Zhenzhong Chen, e-mail: zzchen@ieee.org)
	}
}

\maketitle

\begin{abstract}

Low-light image enhancement (LLIE) aims to improve the illuminance of images due to insufficient light exposure.
Recently, various lightweight learning-based LLIE methods have been proposed to handle the challenges of unfavorable prevailing low contrast, low brightness, etc. In this paper, we have streamlined the architecture of the network to the utmost degree. By utilizing the effective structural re-parameterization technique, a single convolutional layer model (SCLM) is proposed that provides global low-light enhancement as the coarsely enhanced results.
In addition, we introduce a local adaptation module that learns a set of shared parameters to accomplish local illumination correction to address the issue of varied exposure levels in different image regions. Experimental results demonstrate that the proposed method performs favorably against the state-of-the-art LLIE methods in both objective metrics and subjective visual effects. Additionally, our method has fewer parameters and lower inference complexity compared to other learning-based schemes. 

\end{abstract}

\begin{IEEEkeywords}
Low-light image enhancement, convolutional layer, structural re-parameterization
\end{IEEEkeywords}

\section{Introduction}
\IEEEPARstart{D}{ue} 
  to  environment or equipment limitations, images taken under low lighting conditions always result in poor pictures with severe noise, low contrast, and many other problems. Improving the perceptual quality of such low-light images has been a long-standing issue. Traditional solutions include histogram curve adjustment methods\cite{stark2000adaptive,coltuc2006exact,ibrahim2007brightness} and  Retinex-based methods\cite{rahman2004retinex, Naturalness2013, Probabilistic2015}. Although hand-crafted constraints or priors are helpful in improving the quality of the low-light image, the enhanced output always suffers from  over- or under-enhancement in local regions.
 In recent years, with the surge of deep learning, various data-driven methods have been proposed to tackle this problem, including CNN-based method\cite{wei2018deep,DBLP:conf/cvpr/Yang0FW020}, GAN-based method\cite{DBLP:journals/tip/JiangGLCFSYZW21} and flow-based mothed\cite{DBLP:conf/aaai/WangWYLCK22}.
 
  However, although the above mentioned learning-based methods could achieve promising results, most of them require a huge amount of computational resources and long inferring time, making them difficult to be considered for real-time systems or mobile applications. To address this problem, several lightweight models have recently been developed\cite{RUS2021,SCI2022}. They also follow the two typical approaches: curve adjustment-based approach and Retinex-based approach. The representative curve adjustment based methods are the zero-DCE series\cite{2020Zero,2022Zero}. They model low-light image enhancement as a task of image-specific curve estimation. Their network structure only contains several convolution layers to estimate the adjusting curves. As one of the recently proposed Retinex-based methods, SCI\cite{SCI2022} develops a Self-Calibrated Illumination learning framework for fast low-light image enhancement. This method introduces an auxiliary process to
  boost model performance in the training phase and discards the auxiliary structures during inference. SCI has only three convolution layers in the test phase. 
To further push the lightweight LLIE model to the extreme, a single convolution layer model (SCLM) is proposed that could achieve promising results using structural re-parameterization technique.
\begin{figure}[t]
 		\centering
 		\includegraphics[width=8.5cm]{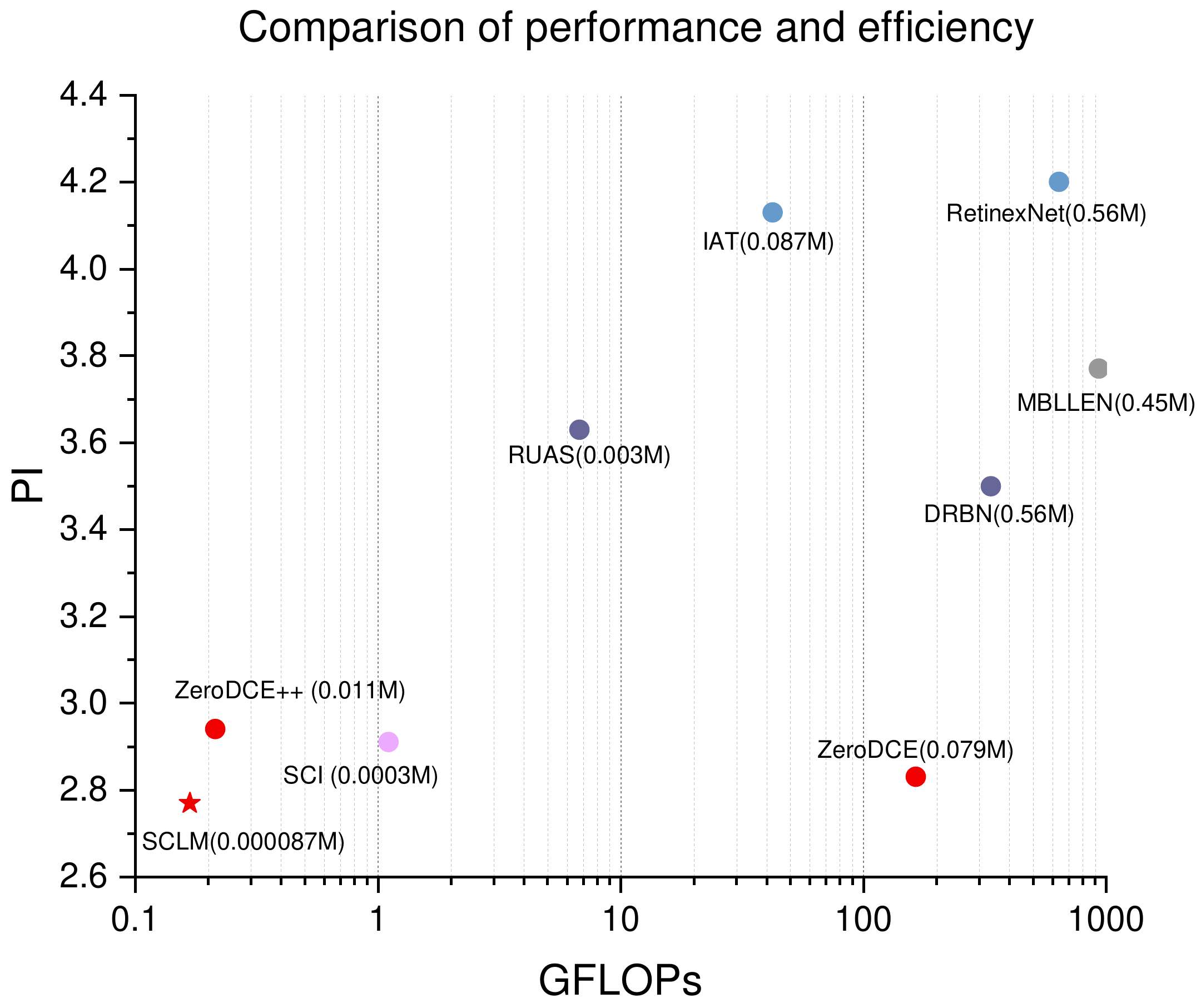}
 		\caption{ Comparison of performance and efficiency. We calculate the average Perceptual Index (PI)$\downarrow$ \cite{DBLP:conf/cvpr/BlauM18,DBLP:journals/cviu/MaYY017,DBLP:journals/spl/MittalSB13} on five real-world datasets. Giga Floating-Point Operations Per Second (GFLOPS)$\downarrow$ is measured on a 1080P (1080$\times$1920) image. The parameter amounts are provided in brackets. Our SCLM (marked with \textcolor{red}{$\star$}) outperforms previous methods in terms of image quality, parameter amount, and GFLOPs.} \label{fig:0}
 \end{figure}
  In addition, due to varying ambient light in the real world, different areas on the same image often have various exposure levels. The enhanced image is thus prone to underexposure in some regions  while overexposure in others. To address this issue, a local adaptation module is introduced that optimizes underexposed and overexposed areas by learning a set of curve adjustment parameters.
   In particular, it is worth noting that this additional module is also extremely lightweight. Instead of employing a complex learnable module to learn the mapping for each pixels  directly,   A set of shared parameters for the adjustment curve are learned to accomplish local exposure correction. Unlike zeroDCE\cite{2020Zero} which performs iterative curve adjustment, curve adjustment is only performed once in SCLM. In short, by jointly optimizing the Retinex-based global enhancement module and  curve-based local adaptation module, SCLM exhibits advanced enhancement performance, and its ultra-lightweight structure design also makes it suitable for edge devices. 
  
As shown in Fig.~\ref{fig:0}, the proposed SCLM outperforms previous methods regarding image quality, parameter amount, and GFLOPs. To the best of our knowledge, our work is the first  to apply the structural re-parameterization technique to the field of LLIE, to build an ultra-simple enhancement structure with one layer of convolution.

 The remainder of the paper is organized as follows: Section
II reviews the related work. Details of the
proposed SCLM are given in
Section III. Experiments and analyses are provided in Section
IV. Finally, the paper is concluded in Section V.

\section{Related Work}
\subsection{Conventional LLIE Methods.}
Traditional LLIE methods can be  divided into two categories: histogram-based methods\cite{coltuc2006exact,ibrahim2007brightness,stark2000adaptive} and
Retinex theroy based methods\cite{Naturalness2013,Probabilistic2015,rahman2004retinex}.
According to Retinex theroy\cite{land1977retinex}, a low-light image can be decoupled into a reflection component and an illumination component. Once the illumination component has been estimated, it can be used to flexibly adjust the exposure level of the image. Specifically, Rahman \emph{et al.} \cite{rahman2004retinex} propose the Multi-Scale Retinex with Color Restoration (MSRCR) algorithm, which transforms low-light images into well-exposured renderings by combining color constancy with local contrast and lightness enhancement. Fu \emph{et al.} \cite{Probabilistic2015} introduce a probabilistic approach to improve the estimation of illumination and reflectance.   Guo \emph{et al.} \cite{LIME2017}  propose a coarse-to-fine method that initially determines the brightness of each pixel by identifying the highest value among the three channels and then enhances the preliminary illumination map by applying a structural constraint to produce the ultimate illumination map.
The histogram-based method attempts to redistribute the
luminous intensity on the histogram globally or locally. Coltuc \emph{et al.}\cite{coltuc2006exact} regard the low-light image enhancement problem as a K-dimensional space optimization problem with ordering strict among pixels. Haidi \emph{et al.} \cite{ibrahim2007brightness} propose a brightness-preserving dynamic histogram equalization method, thus fulfilling the requirement of maintaining the mean
brightness of the input and the enhanced image. Lee \emph{et al.} \cite{lee2013contrast} 
present a contrast enhancement algorithm that amplifies the gray-level differences between adjacent pixels in a tree-like layered structure.
\subsection{Learning-based LLIE Methods.}
In recent years, data-driven methods for low-light image enhancement have received increasing attention. Unlike other restoration tasks such as super-resolution and denoising, some work\cite{survey2022} points out that combining traditional methods and deep learning could achieve better performance than direct end-to-end enhancement due to their physical interpretability. Similar to the traditional technical route, the mainstream learning-based method can also be divided into Retinex-based methods\cite{shen2017msr,wei2018deep, RUS2021, SCI2022} and curve adjustment-based methods\cite{2020Zero,2022Zero}. In practice, Retinex-based deep learning methods generally concatenate three channels of color images and then use a neural network to predict the illuminance component.     Shen \emph{et al.} \cite{shen2017msr} establish a relationship between multi-scale Retinex and the feedforward convolutional
neural network to enhance the low-light image. Wei \emph{et al.} \cite{wei2018deep} build a LOw-Light dataset (LOL) containing low/normal-light image pairs and propose a deep Retinex-Net to learn consistent reflectance shared by paired low/normal-light images. RUAS\cite{RUS2021} first develops models to represent the inherent underexposed structure of low-light images and then unfolds optimization processes to construct the comprehensive propagation structure for LLIE. Zhao \emph{et al.}\cite{DBLP:journals/tcsv/ZhaoXWOYK22} propose a generative strategy for
Retinex decomposition. They build a unified deep
framework to estimate the latent illuminance component and
to perform low-light image enhancement. Fan \emph{et al.} \cite{DBLP:journals/tcsv/FanFGCC22} propose a low-light image enhancement model to address the problem of uneven exposure or partial overexposure with illumination constraint. Most recently, SCI\cite{SCI2022} constructs a self-calibrated module that employs an Auxiliary block to realize the convergence between each stage in the training phase and only uses the single basic block for inference. 
Some other representative methods are based on the curve adjustment scheme. ZeroDCE\cite{2020Zero} formulates light enhancement as a task of image-specific curve adaptation
with a deep network. Based on \cite{2020Zero}, ZeroDCE++\cite{2022Zero} further approximate pixel-wise and higher-order curves by iteratively applying itself and also discuss multiple options to balance the enhancement
performance and the computation cost of resources for high-resolution images. There are also some end-to-end approaches\cite{DBLP:journals/tcsv/LiFH21,DBLP:journals/tcsv/LiuWW22} that directly enhance low-light images with a deep model. 
Although they achieve pleasing results, their high computational complexity and long inference times make them difficult to deploy in practice.

 \begin{figure*}[t]
	\centering
	\includegraphics[width=17.5cm]{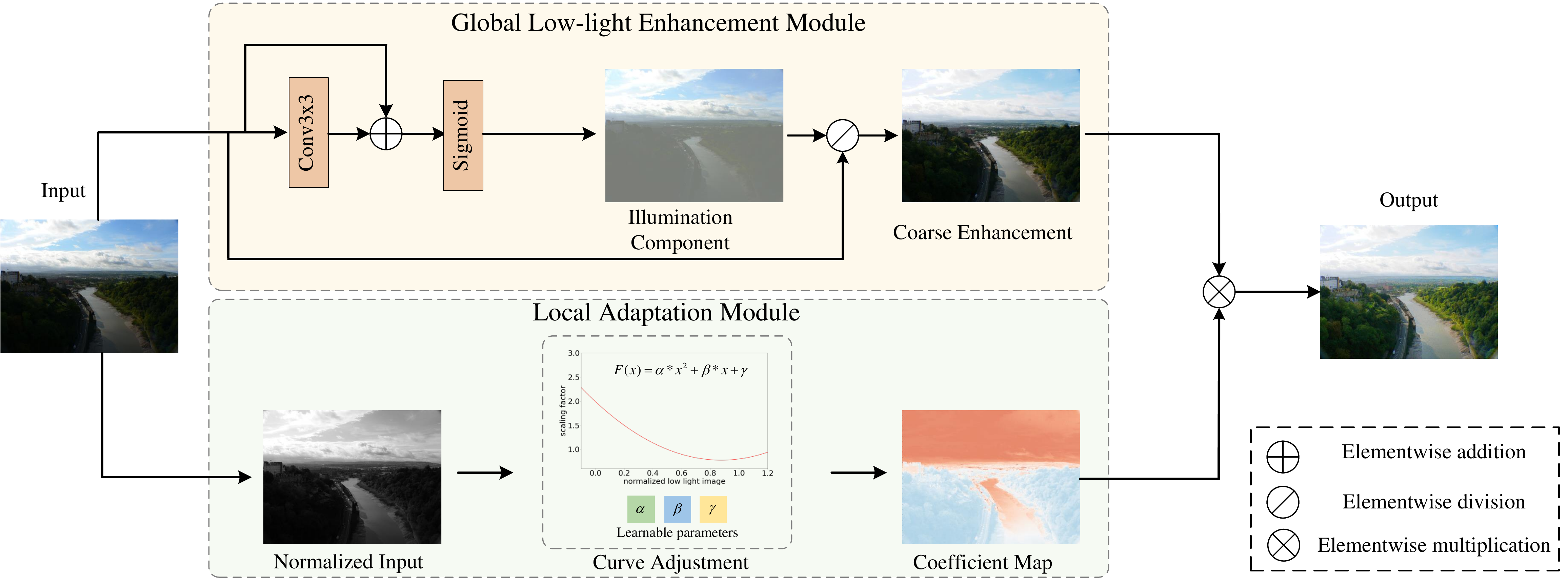}
	\caption{The framework of the proposed SCLM. It consists of two parts, global enhancement, and local illuminance adaptation. The global enhancement module is a typical Retinex-based design, where a low-light image is first fed into a network with only one convolutional layer (during inference) to estimate the illumination map of the image. Then, the original image is divided by the illumination map to obtain the coarsely enhanced result. Meanwhile, the normalized illumination map is fed into a quadratic function curve which is driven by learnable parameters to obtain a local adaptation map. Finally, the coarsely enhanced result is element-wise multiplied with the local adjustment map, outputting the final enhanced image.  
	} \label{fig:1}
\end{figure*}

\section{The proposed  SCLM method}
 Inspired by the above pioneering work, our approach consists of two enhancement steps from coarse to fine, which can be viewed as combining the Retinex-based approach and the curve estimation approach.
Before delving into the specific designs, let's first briefly describe the pipeline of the proposed method. The overview of the proposed method is given in Fig.~\ref{fig:1}. SCLM is constituted of two parts, the global enhancement module and the local adaptation module. Specifically, The global enhancement module contains only one layer of convolution at the inference time to estimate the illumination component. An enhanced image can be obtained by the original low-light image dividing with the estimated illumination part according to the Retinex theory. Since different areas of an image have diverse illuminations, a learnable quadratic curve is then jointly optimized through fine adjustment for different image regions. Instead of using complex iterative adjustments for each pixel\cite{2020Zero}, our curve adjustment process is performed only once, and all pixels share the same set of curve adjustment parameters.	\vspace{-1em}

 \subsection{Global low light enhancement}
 \subsubsection{Plain Structure}
The global low-light enhancement (GLLE) module is built upon the Retinex theroy\cite{land1977retinex} :
 \begin{equation}
 	\begin{aligned}
 	 y = z \otimes x
 	\end{aligned}
 	\label{f1}
 \end{equation}
where $x$  represents 
illumination component, $y$ represents  low-light observation, and  $z$ denotes  desired clear image.

Generally speaking, the luminance component is considered as the key part of optimization. Therefore, many  Retinex-based methods focus on  estimating the luminance component more accurately and efficiently. Recently, many ultra-lightweight deep learning models have been proposed, continuously simplifying the structure of network, among which SCI\cite{SCI2022} comprises only three convolutional layers.

To further push the extreme of structure simplicity for the sake of decreasing computation and complexity. A simple model with only one convolutional layer is first constructed, which is refered as the plain model.

 \begin{figure}[t]
	\centering
	\includegraphics[width=8.5cm]{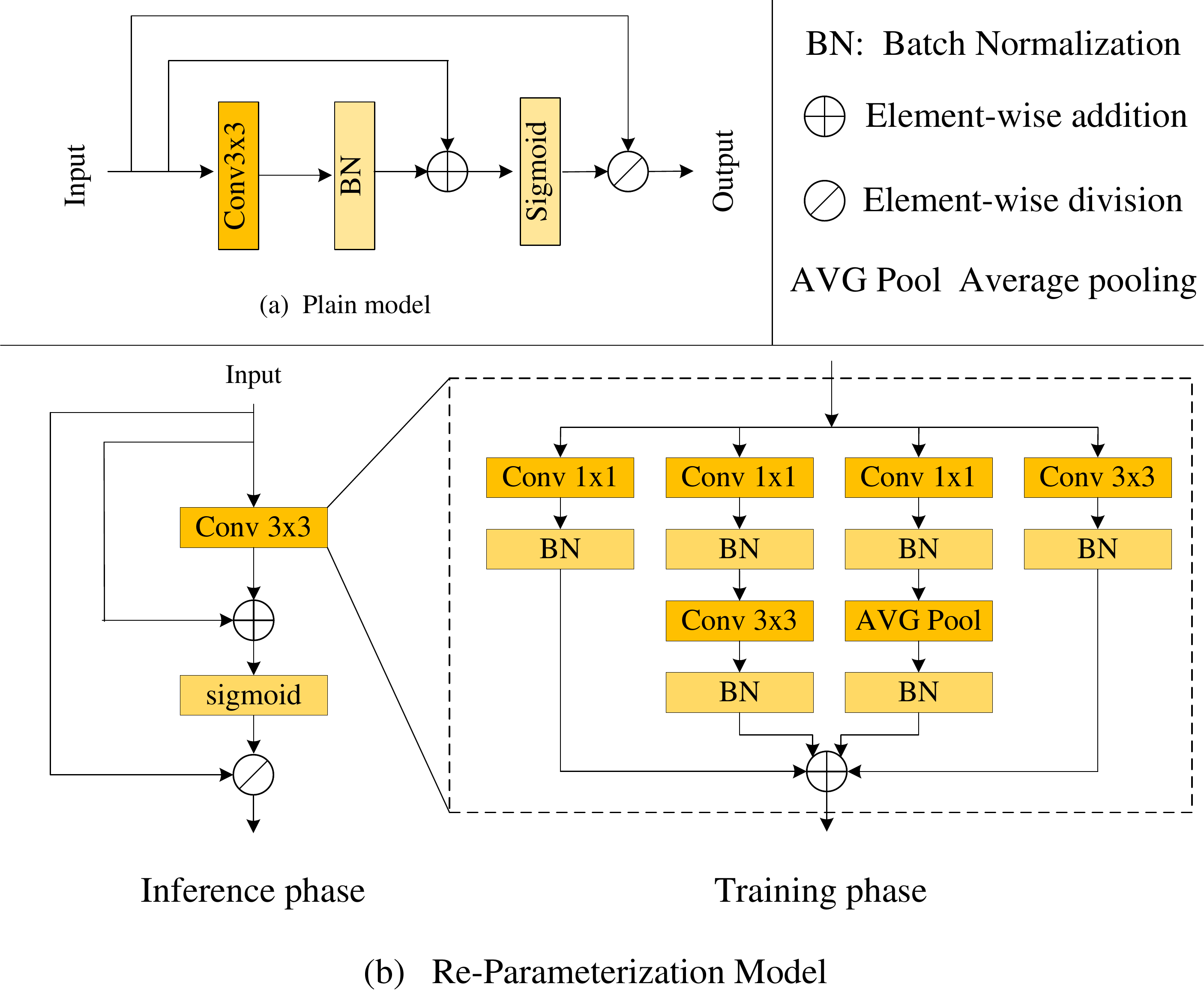}
	\caption{Illustrations of the one-layer model.
	 (a) The plain structure, which only consists of a 3$\times$3 convolution layer followed by a batch normalization. (b) A multi-branch structure that contains  parallel convolutions and pooling operations in training phase, which can be collapsed into a plain 3$\times$3 convolution during inference.}\label{fig:2} 
\end{figure}

 As shown in Fig.~\ref{fig:2} (a), the plain model consists of only one convolutional layer followed by a batch normalization layer. A residual connection is employed to improve the stability of training and a sigmoid activation function is used to ensure that the estimated luminance component falls between 0 and 1. The training process is carried out in a fully supervised manner. To be specific, the experiments are conducted on the Multi-Exposure dataset\cite{DBLP:conf/cvpr/AfifiDOB21}, which contains image pairs with different exposure levels. Here, we  use the $\mathcal{L}_{1}$ loss function and supervise the model using pairs of low-light inputs and corresponding enlightened images.
 \begin{figure*}[t]
	\centering
	\includegraphics[width=17.5cm]{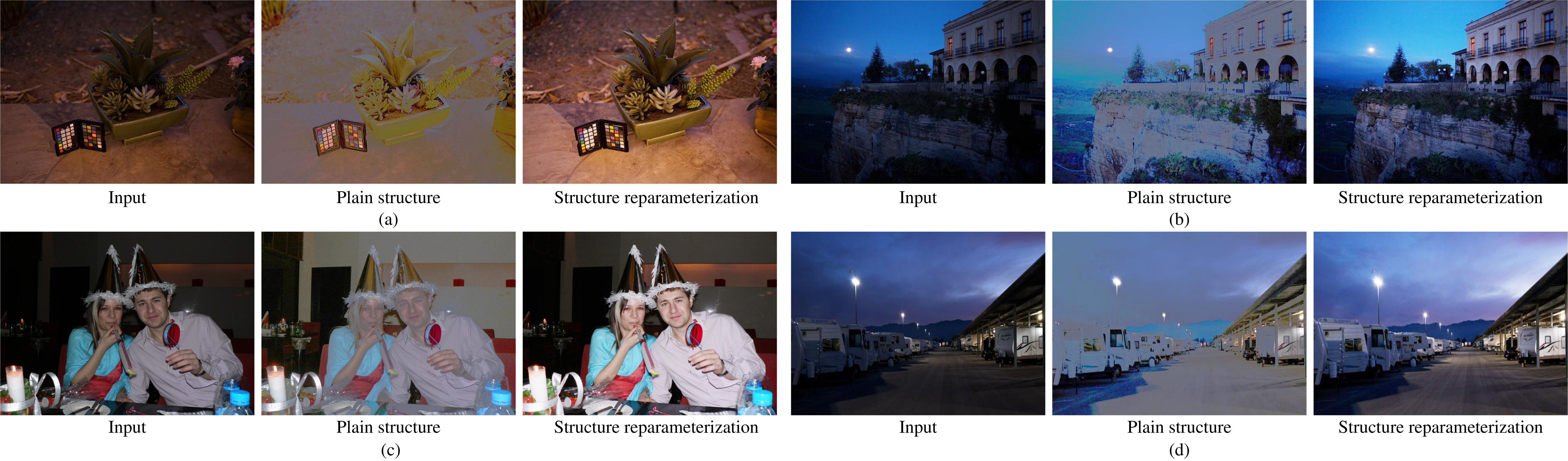}
	\caption{Enhancement results of different model topology. Plain structure appears severe color unsaturation artifacts, while model with structural re-parameterization can produce much decent results. Zoom in for a better view.
	}\label{fig:3} 
\end{figure*}
The enhancement results for the model with plain structure are shown in Fig.~\ref{fig:3}. As can be seen, a vanilla one-layer model cannot achieve  pleasing perceptual effects. For example, in Fig~\ref{fig:3} (a) (b), there exists apparent color distortion in the overall image, and there are significant color unsaturation and artifacts in Fig~\ref{fig:3} (c) (e.g., face region) and (d) (e.g., sky region). This experiment indicates the representative capability  of the plain structure is limited. 

  \subsubsection{ Structural Re-parameterization }
  Since the plain single-layer model cannot produce satisfactory results, a natural solution is to deepen the network with sequence layers or adopt a multi-branch structure. However, increasing the depth or width of the network will inevitably lead to increased computational complexity. To increase the representational capability of the model as much as possible while ensuring its simple design, the structural re-parameterization technique is introduced. 
  Structural re-parameterization \cite{DBLP:conf/iccv/DingGDH19,DBLP:conf/cvpr/Ding0MHD021,DBLP:conf/cvpr/Ding0HD21} refers to utilizing a multi-branch structure during training to enrich the feature space and then transforming the model into an equivalent single-branch structure during inference to enhance the representational capability of a single convolution. Recently, some low-level tasks, such as super-resolution\cite{DBLP:conf/mm/ZhangZZ21,DBLP:conf/mm/WangDS22}, have utilized structural re-parameterization techniques to improve the performance of edge-oriented methods.

Inspired by these previous methods, we design a multi-branch model incorporating various convolution and pooling operations to accomplish the estimation of the luminance component.
 As shown in Fig.~\ref{fig:2} (b), our multi-branch GLLE  consists of four parallel branches including a 3$\times$3 convolution branch, a 1$\times$1 convolution branch, a sequence of 1$\times$1 convolutions followed by 3$\times$3 convolutions branch, and a sequence of 1$\times$1 convolutions followed by an average pooling branch.
We next introduce the transformations in our model, enabling us to collapse the multiple-layer and multiple-branch model into a single convolution.\\
\textbf{Merge BN into preceding Conv. }
A common optimization strategy for modern deep learning framework is to merge the batch normalization (BN) into its preceding convolution at inference time to accelerate the speed. This process can be depicted as follows.
Let $\mu$ and $\sigma$ denote the accumulated channel-wise mean and standard deviation, $\gamma$ and $\beta$ denote the learnable scaling factor and bias. Then the output feature $O$ can be obtained  by:
\begin{equation}
	\begin{aligned}
	{O} &= ({I} \circledast {W} - \mu).\frac{\gamma}{\sigma}+ \beta,
	\end{aligned}
\end{equation}
where $I$ represents the input feature and $W$ represents the weight of the kernel, $\circledast$ denotes convolution operator.
According to the homogeneity of the convolutional layer, BN can be fused into its preceding Conv for inference by building a new single Conv layer with kernel $W^{\prime}$ and bias 
$b^{\prime}$ and their values are assigned as:
\begin{equation}
	\begin{aligned}
		W^{\prime} \leftarrow \frac{\gamma}{\sigma}.W  \quad \quad \quad
		b^{\prime} \leftarrow  - \frac{\mu \gamma}{\sigma}.W  + \beta.
	\end{aligned}
\end{equation}
After the training of the network, 
the transformed parameters (kernel and bias) can be saved  for model inference.  \\
\textbf{Merge sequential Convs.}
According to the associativity of convolution, the sequence of 1$\times$1 Conv - 3$\times$3 Conv can be merged into one single 3$\times$3 Conv:
 \begin{equation}
	\begin{aligned}
		O &= (I \circledast W_{1} + b_{1}) \circledast W_{3} + b_{3} 	\\	
		& = I \circledast \underbrace{({W}_1 \circledast {W}_3)}_{\text{merged weight}} + \underbrace{({b}_1 \circledast {W}_3 + {b}_3)}_{\text{merged bias}} 	
	\end{aligned}
\end{equation}
where $W_1$ and $W_3$ are the weight of the original $1\times1$ and $3\times3$ Conv, $b_1$ and $b_3$ are the  bias of 1$\times$1 and 3$\times$3 Conv.  For the sequence of 1$\times$1 Conv-AVG Pooling branch,  the pooling operation can be regarded as a convolution kernel with pre-specified parameters. Precisely,  an average pooling with kernel size $K$ and stride $S$ can be replaced by a Conv with the same $K$ and $S$, and the kernel values are set as $\frac{1}{K^{2}}$. By utilizing this re-parameter technique, the sequential Convs are shrinked into a single convolutional layer.  \\
\textbf{Merge multiple branch Conv.}
Since the convolution operation is a linear transformation with additivity, two or more parallel Conv layers with the same kernel size can be merged into a single Conv layer. For two convolution kernels with different kernel sizes (e.g., 3$\times$3 convolution and 1$\times$1 convolution), they can be unified into the same size by padding the smaller convolution kernels with zeros. 
Then, The merge process of $n$ parallel convolution kernels can be formulated  as:
\begin{equation}
	\begin{aligned}
		W^{\prime} &\leftarrow  W^{(1)} + W^{(2)}+...+W^{(n)},\\
		b^{\prime} &\leftarrow  b^{(1)} + b^{(2)}+...+ b^{(n)}
	\end{aligned}
\end{equation}
where $ W^{(i)}$ stands for the weight of the $i^{th}$ kernel and $ b^{i}$ denotes the weight of the $i^{th}$ bias, $W^{\prime}$ and $b^{\prime}$ denotes the merged weight and bias.

By employing these steps of transformations,   diverse branches of GLLE can be combined into a single convolution, while the representative capability is maintained. Through  experiments, we verify that with the support of structural re-parameterization, significant improvements can be achieved in the restoration results. And its performance is comparable to many methods in terms of visual effects whose parameter counts thousands of times higher. As can be seen in Fig.~\ref{fig:3}, the model with structure reparameterization performs significantly better than the plain structure.
Through experiments, it is verified that  considerable results can be achieved with very few learnable parameters, indicating that even elementary structures have great potential for LLIE task.

\subsection{Local Adaptation Module}

Although a single convolutional layer can achieve acceptable results, its ability to enhance images under complex low-light conditions needs further improvement. Similar to many Retinex-based methods\cite{wei2018deep, RUS2021, SCI2022}, it is extremely challenging to process images containing different exposure levels in a single image. It is easy for Retinex-based methods to encounter situations where some areas are overexposed while other blacklit regions are insufficiently enhanced.

This is due to the reason that methods based on Retinex theory often restrict the illumination map to be within 0 to 1. The enhancement process of dividing the low-light input image by the illumination map  can  thus only  brighten each pixels. In contrast, methods based on curve adjustment are relatively more flexible. Inspired by lighting enhancement \cite{2020Zero} and automatic exposure correction methods\cite{yuan2012automatic}, a curve adjustment module is designed for SCLM  to finely enhance the image further.

This local adaptation module will learn a set of curve adjustment parameters in a data-driven manner. The curve can be used to enhance dark areas while suppressing overexposure in bright areas. In addition, the curve should have the following characteristics: (1) it should be simple enough and learnable by backpropagation.
(2) as different areas require different degrees of enhancement, the curve should satisfy a non-linear mapping relationship. Taking into account these factors and the experience of previous work\cite{2020Zero,2022Zero}, a quadratic function is used to model this mapping relationship.

For each input RGB image, it is first converted to the YCbCr color space and then the difference between the Y channel  of each pixel and the median Y channel of the entire image is calculated:
 \begin{equation}
	\begin{aligned}
		I_{y} &= RGB2Ycbcr(I_{RGB}) \\
		\hat{\mathcal{M}} &= I_{y} - median(I_{y})
	\end{aligned}
	\label{f2}
\end{equation}
where $median$  refers to taking the median value of the image and $\hat{\mathcal{M}}$ denotes a condition map.

  The condition map is further normalized in range (0,1) to ensure training stability:
 \begin{equation}
	\begin{aligned}
	\overline{ \mathcal{M}} = \frac{\hat{\mathcal{M}} - Min(\hat{\mathcal{M}} )}{ Max(\hat{\mathcal{M}}) - Min(\hat{\mathcal{M}})}
	\end{aligned}
	\label{f3}
\end{equation}
where $\overline{ \mathcal{M}}$ denotes the normalized condition map.

In order to reduce the impact of local outlier pixels and introduce more non-linearity, maximum pooling operation is applied to the conditional feature map, which can also ensure the local smoothness of the conditional feature map:
 \begin{equation}
	\begin{aligned}
		 \mathcal{M} = MaxPool(\overline{ \mathcal{M}})
	\end{aligned}
	\label{f4}
\end{equation}

Afterward, the conditional map $\mathcal{M}$ is used to the quadratic function to compute the modulation coefficients for each pixel. We assume that the input image contains $N$ pixels, and the modulation process can be formulated  as:
 \begin{equation}
	\begin{aligned}
		C_{i} = \alpha*x_{i}^{2}+\beta*x_{i} + \gamma ,\\
		where \quad i \in (1,N) , \quad x_{i}  \in \mathcal{M}.
	\end{aligned}
	\label{f5}
\end{equation}
Here, $\alpha$, $\beta$, and $\gamma$ are three learnable parameters that are shared across all pixel points. In other words, there are only three learnable parameters needed for the entire image.

\subsection{Output Fusion}
Finally, we multiply the Coarse Enhanced image $I_{coarse}$ with the modulated coefficient map $C$ to obtain the enhanced result:
 \begin{equation}
	\begin{aligned}
	  I_{en} = C \otimes I_{coarse},
	\end{aligned}
	\label{f6}
\end{equation}
where $\otimes$ denotes element-wise multiplication. During multiplication, some clamp operations are applied to prevent out-of bounds pixel values. So far, the enhancement process of the one layer model has been completed. It is worth noticing that the global low-light enhancement module and local adaptation module are jointly optimized in an end-to-end manner.  The entire model has only 87 learnable parameters, including a 3$\times$3 convolution (84 parameters) and three coefficients of a quadratic function (3 parameters). The specific effectiveness of each component and corresponding analysis will be provided in the experimental section.

	\vspace{1em}
\section{Experimental Results}
\begin{table*}[t]
	\caption{ Perceptual index (PI) $\downarrow$ / naturalness image quality evaluator (NIQE)  $\downarrow$  on five real world low-light  datasets The best results are highlighted with {\color{red} red} color, and the second best results are  highlighted with {\color{blue} blue} color}
	\centering
	\resizebox{0.85\linewidth}{!}{
		\begin{tabular}{c|c|c|c|c|c|c|c}
			\hline
			Type                                                            & Method               & LIME      & VV        & MEF       & NPE       & DICM      & Average   \\ \hline
			\multicolumn{1}{l|}{\begin{tabular}[c]{@{}l@{}}Traditional method\end{tabular}} & LIME\cite{LIME2017}           & 2.88/{\color{blue}{3.99}} & {\color{blue}{2.88}}/{\color{red}{2.51}} & {\color{red}{2.43}}{\color{blue}{/3.43}} & 3.10/4.26 & {\color{blue}{3.08}}/{\color{blue}{3.78}} & 2.88/3.59 \\ \hline
			\multirow{3}{*}{\begin{tabular}[c]{@{}c@{}}Supervised\\ learning\end{tabular}}       & RetinexNet\cite{wei2018deep}     & 4.65/4.62 & 3.56/2.62 & 4.30/4.32 & 4.23/4.59 & 4.25/4.42 & 4.20/4.11 \\
			& MBLLEN\cite{DBLP:conf/bmvc/LvLWL18}         & 3.78/4.52 & 3.39/4.16 & 3.76/4.78 & 4.01/4.46 & 3.89/4.16 & 3.77/4.42 \\
			& DRBN\cite{DBLP:conf/cvpr/Yang0FW020}       & 3.52/4.73 & 3.57/3.22 & 3.32/4.24 & 3.36/4.43 & 3.74/3.93 & 3.50/4.11 \\
			& IAT\cite{DBLP:conf/bmvc/CuiL0SG00H22}       & 4.31/4.91 & 4.40/4.06 & 3.68/4.29 & 4.02/4.51 & 4.22/4.39 & 4.13/4.43 \\ \hline
			\multirow{6}{*}{\begin{tabular}[c]{@{}c@{}} Zero reference\\ learning\end{tabular}}  & RUAS\cite{RUS2021}            & 3.13/4.42 & 3.92/4.48 & 2.84/3.91 & 4.04/9.22 & 4.20/4.85 & 3.63/5.37 \\
			& SCI\_easy\cite{SCI2022}      & 2.96/4.23 & 3.24/3.06 & 2.80/3.74 & 2.88/4.02 & 3.33/3.83 & 3.04/3.78 \\
			& SCI\_medium\cite{SCI2022}    & 3.06/4.29 & 3.19/3.02 & 2.66/3.72 & 3.45/4.49 & 3.63/4.24 & 3.20/3.95 \\
			& SCI\_difficult\cite{SCI2022} & 2.87/4.12 & 2.96/2.84 & 2.56/3.70 & 2.97/4.14 & 3.17/4.17 & 2.91/3.79 \\
			& ZeroDCE\cite{2020Zero}        & {\color{red}{2.80}}/{\color{red}{3.88}} & 2.95/2.62 & {\color{blue}{2.47}}/{\color{red}{3.40}} & {\color{blue}{2.81}}/{\color{blue}{3.89}} & 3.11/3.85 & {\color{blue}{2.83}}/{\color{red}{3.53}} \\
			& ZeroDCE++\cite{2022Zero}      & 2.94/4.04 & 2.99/{\color{blue}{2.61}} & 2.54/3.47 & 2.91/3.97 & 3.34/3.93 & 2.94/3.60 \\ \hline
			Supervised learning                                                  & 
			SCLM (Ours)           & {\color{blue}{2.85}}/4.06 & {\color{red}{2.86}}/2.65 & 2.58/3.57 & {\color{red}{2.75}}/{\color{red}{3.77}} & {\color{red}{2.83}}/{\color{red}{3.69}} & {\color{red}{2.77}}/{\color{blue}{3.55}} \\ \hline
	\end{tabular}}\label{tab:1}
\end{table*}
\begin{figure*}[htbp]
	
	\centering
	\includegraphics[width=16.5cm,height=8cm]{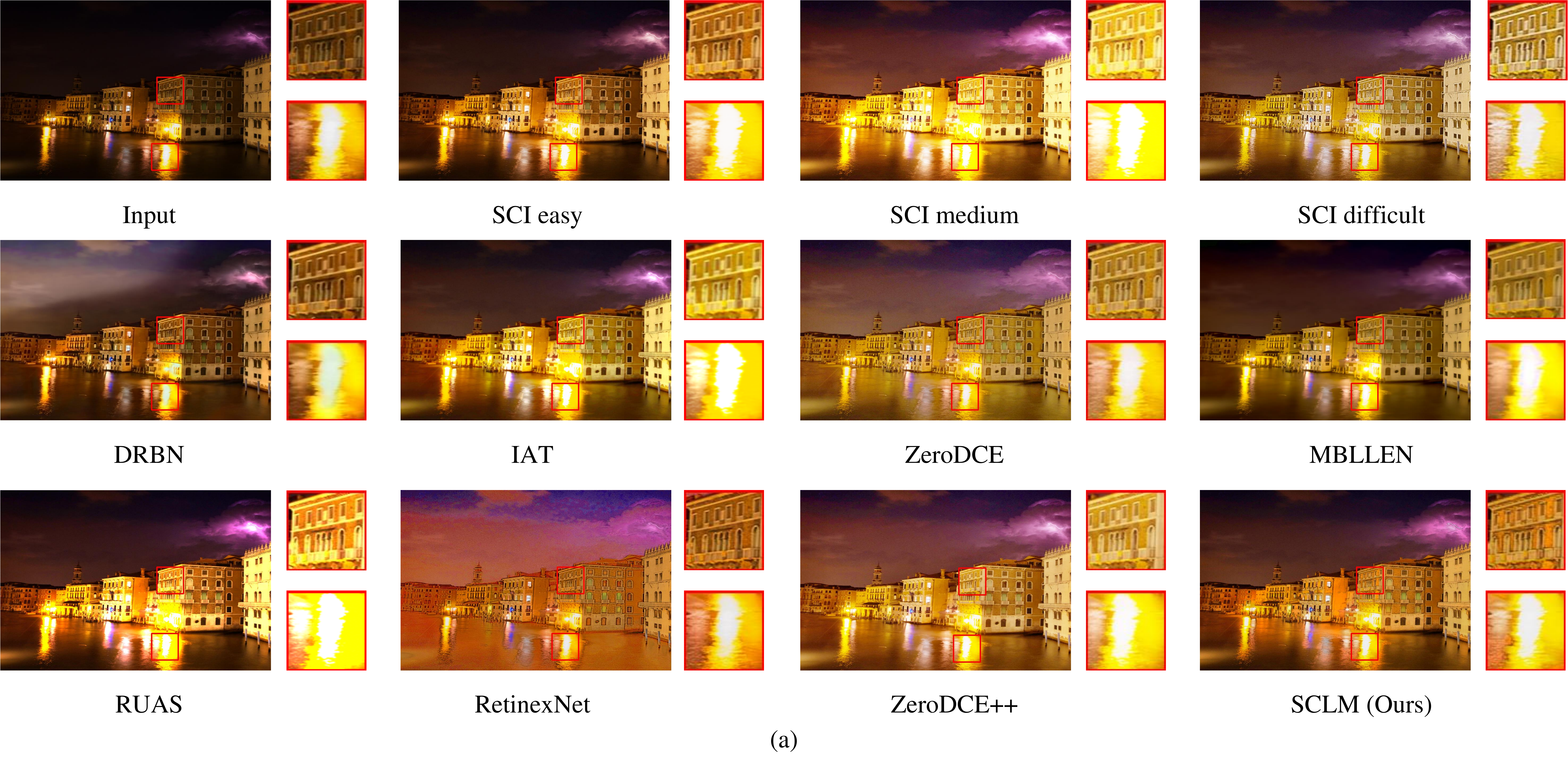}
\end{figure*}
\begin{figure*}[!t]
	\vspace{-1em}
	\centering
	\includegraphics[width=16.5cm,height=8cm]{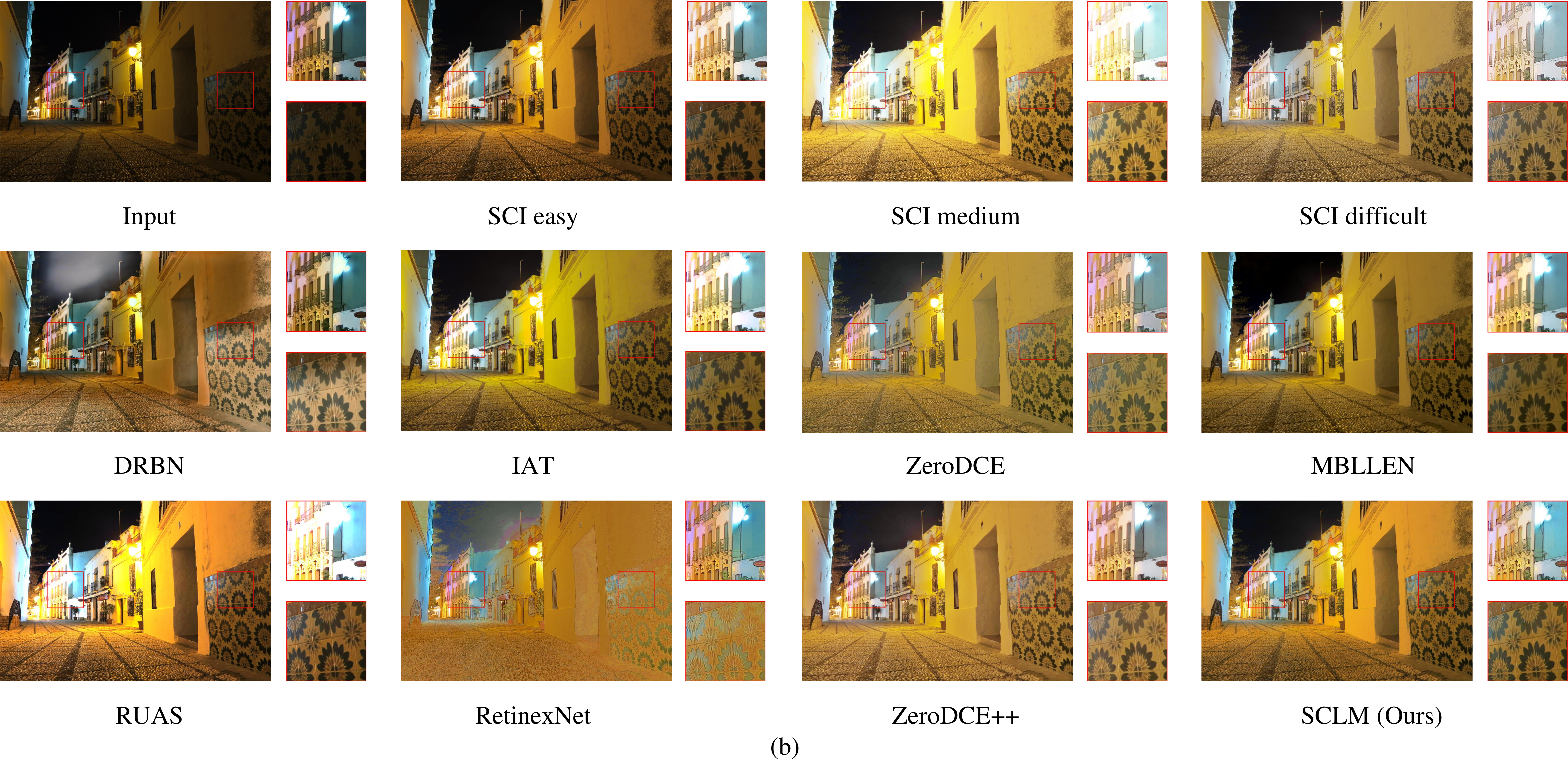}
	\caption{ Visual comparison of different methods on LIME\cite{LIME2017}. Parts of areas are zoomed in with red boxes.} \label{fig:LIME_2}
\end{figure*}
\begin{figure*}[htbp]
	\vspace{-1em}
	\centering
	\includegraphics[width=16.5cm,height=8cm]{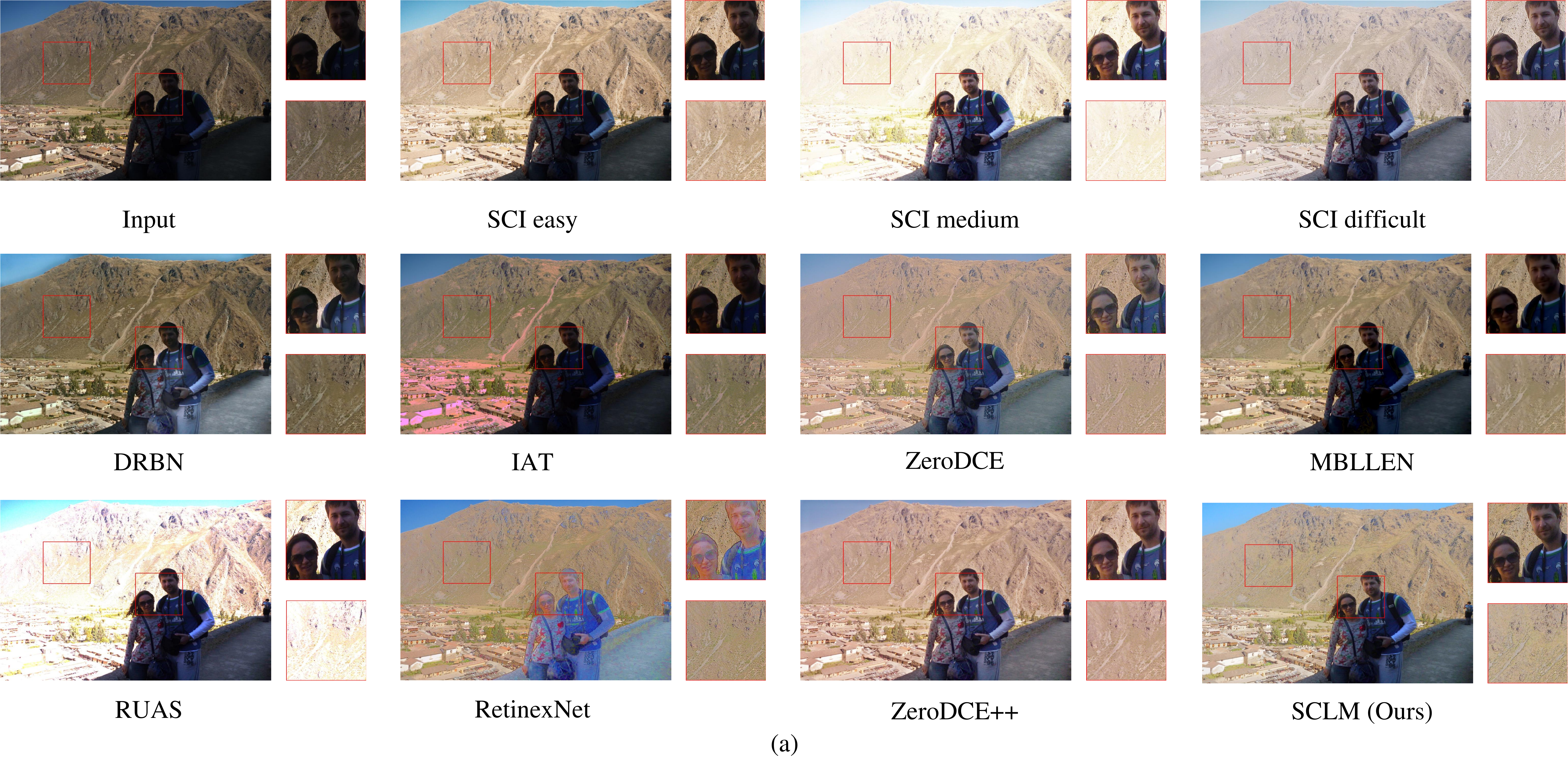}
\end{figure*}
\begin{figure*}[htbp]
	\vspace{-1em}
	\centering
	\includegraphics[width=16.5cm,height=8cm]{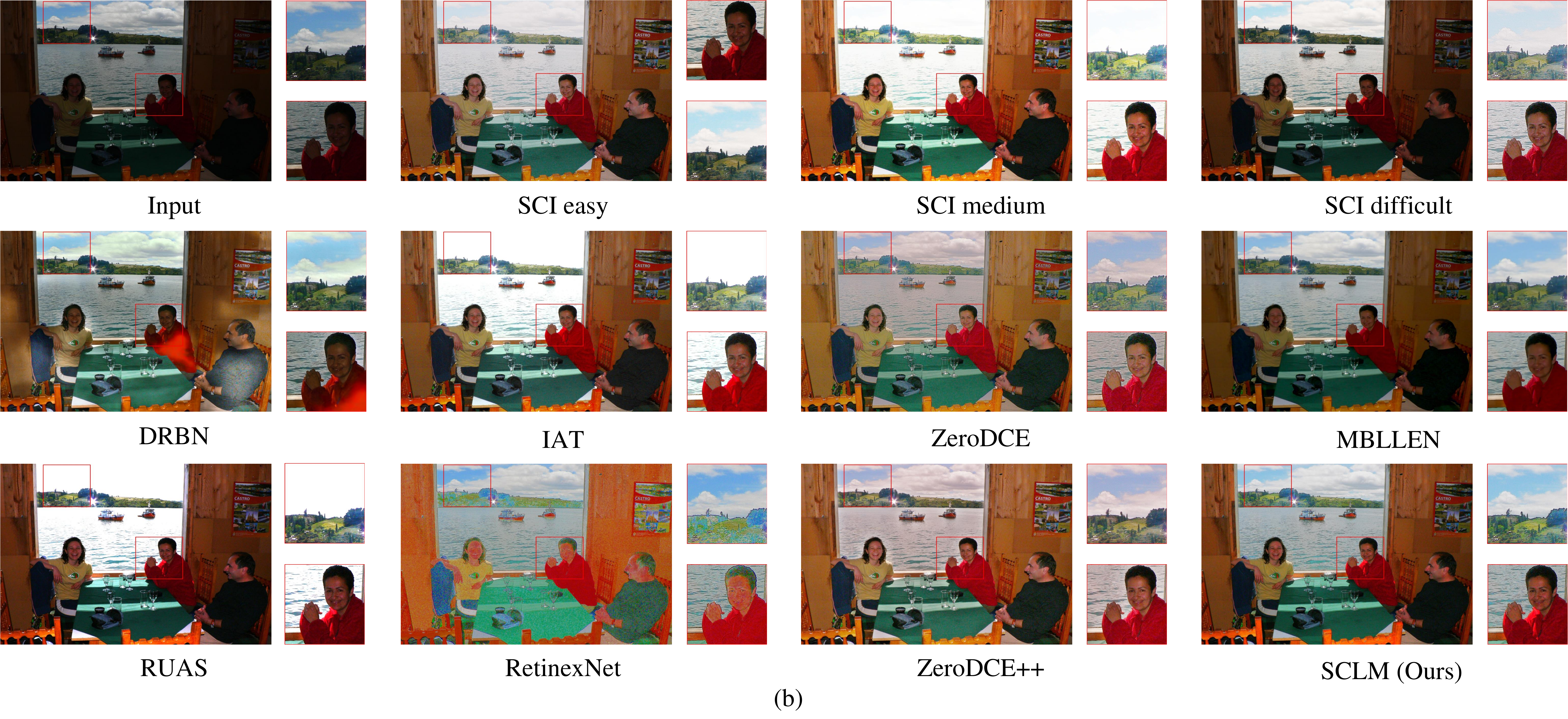}
	\caption{ Visual comparison of different methods on VV\cite{vonikakis2021busting}. Parts of areas are zoomed in with red boxes.} \label{fig:VV_2}
\end{figure*}
\begin{figure*}[htbp]
	\vspace{-1em}
	\centering
	\includegraphics[width=16.5cm,height=8cm]{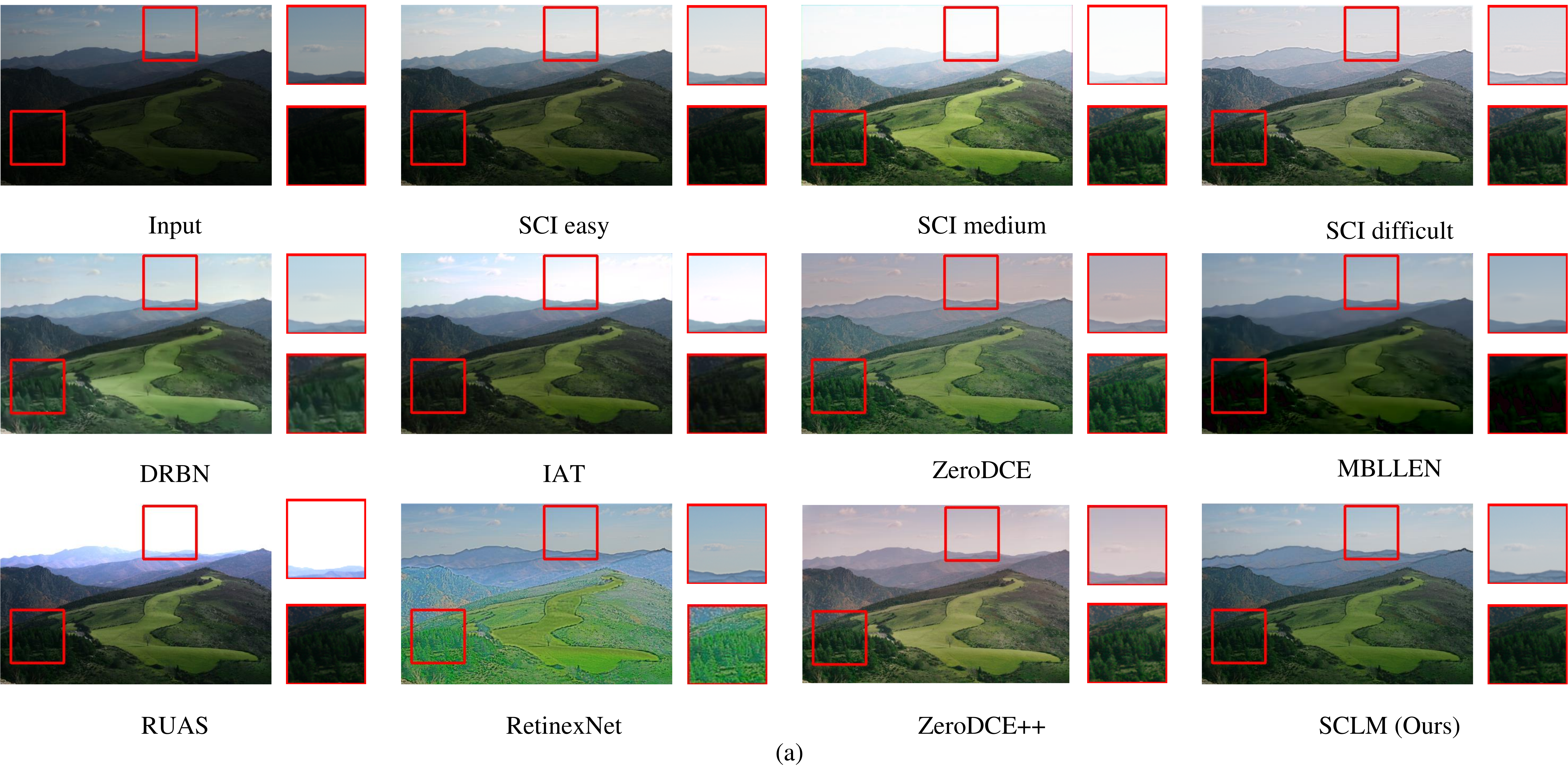}
\end{figure*}
\begin{figure*}[htbp]
	\vspace{-1em}
	\centering
	\includegraphics[width=16.5cm,height=8cm]{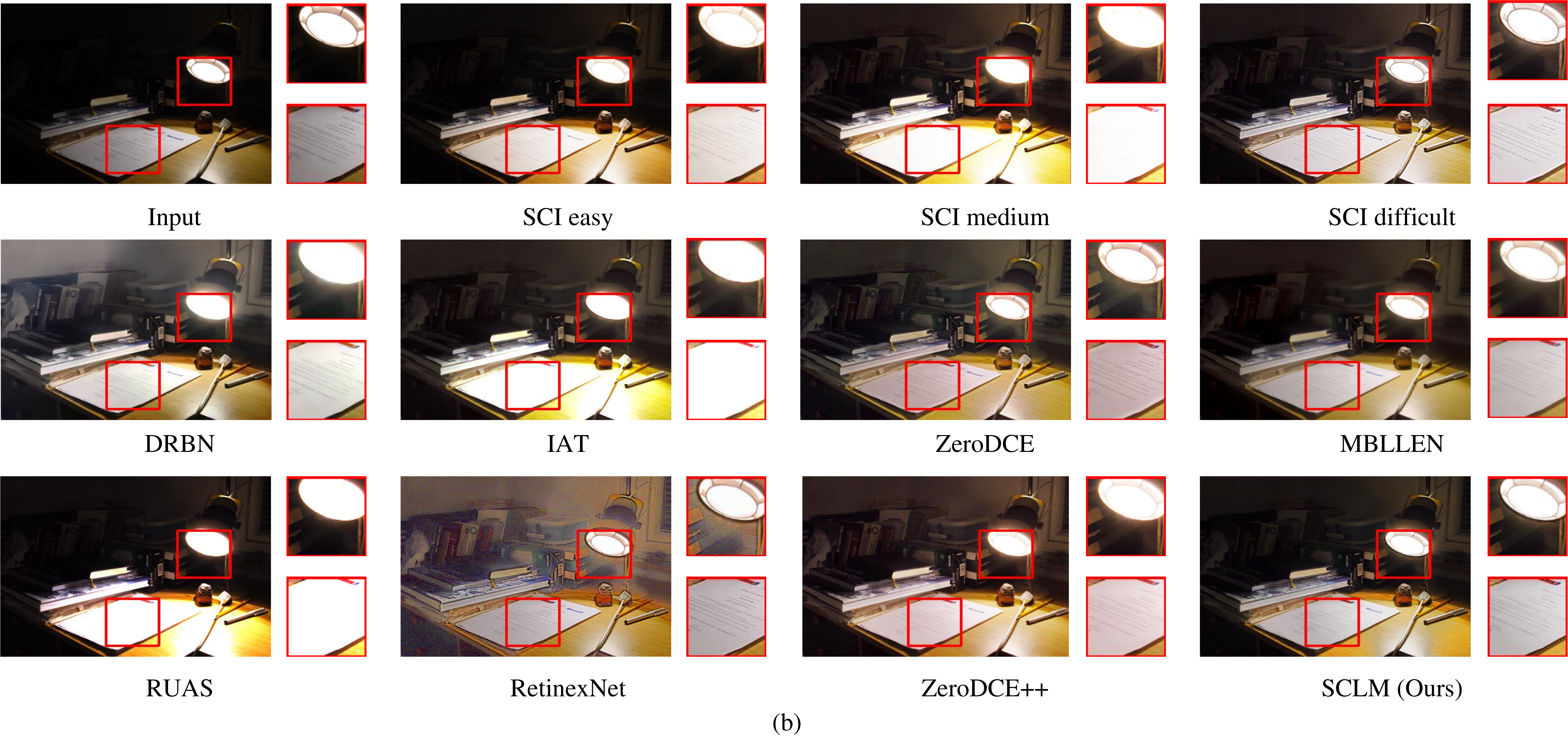}
	\caption{ Visual comparison of different methods on MEF\cite{DBLP:journals/tip/MaZW15}. Parts of areas are zoomed in with red boxes.} \label{fig:MEF_2}
\end{figure*}
\begin{figure*}[htbp]
	\vspace{-1em}
	\centering
	\includegraphics[width=16.5cm,height=8cm]{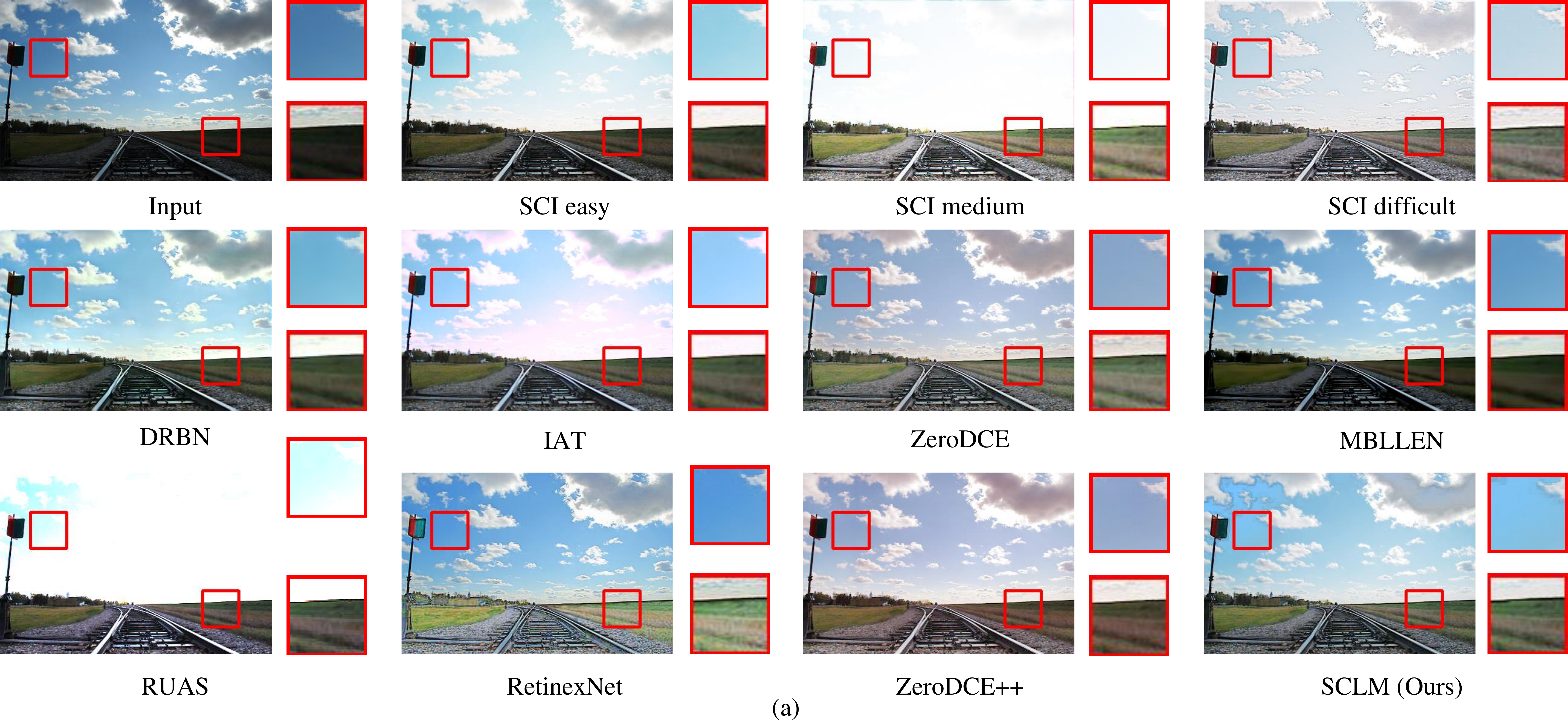}
\end{figure*}
\begin{figure*}[htbp]
	\vspace{-1em}
	\centering
	\includegraphics[width=16.5cm,height=8cm]{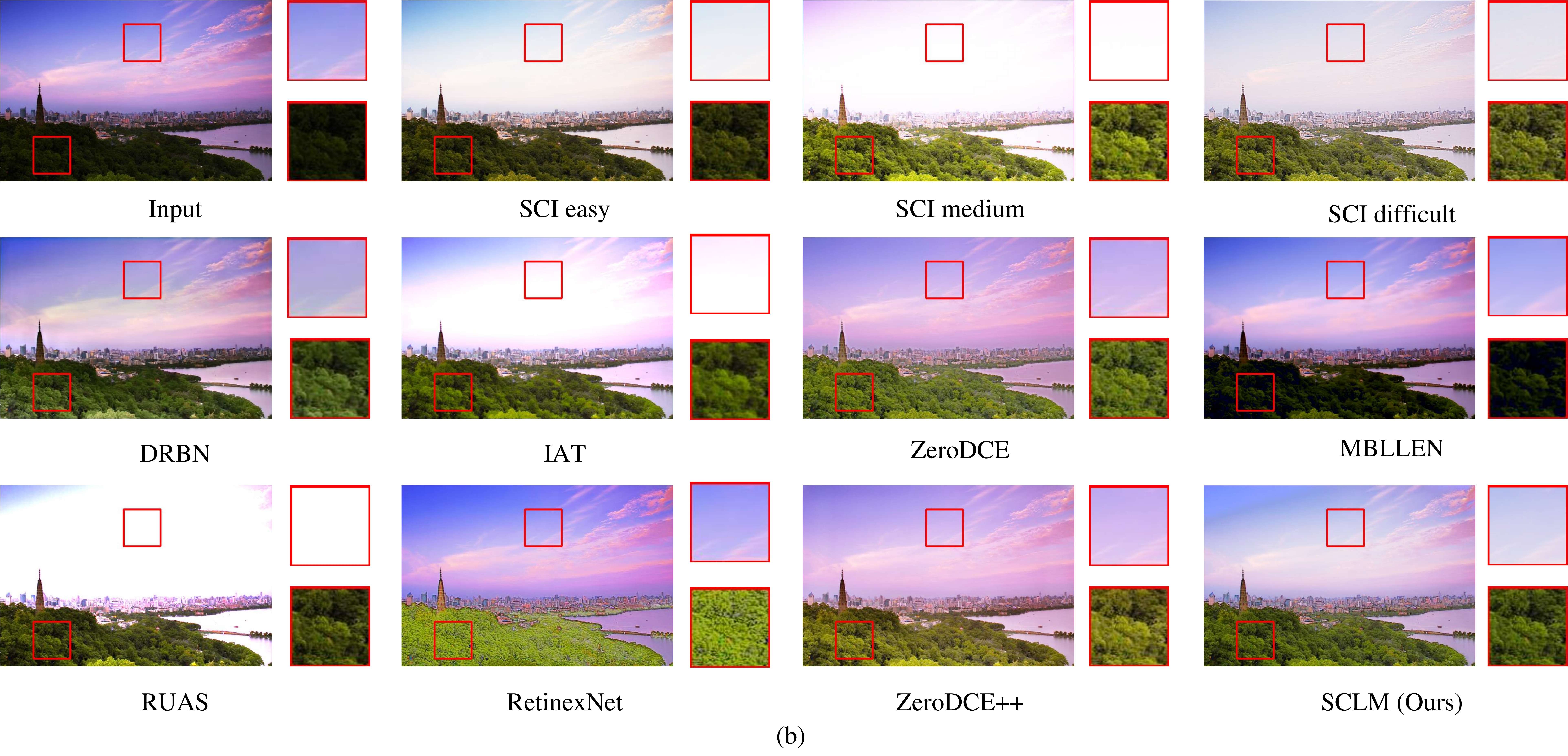}
	\caption{ Visual comparison of different methods on NPE\cite{Naturalness2013}. Parts of areas are zoomed in with red boxes.} \label{fig:NPE_2}
\end{figure*}
\begin{figure*}[htbp]
	\centering
	\includegraphics[width=16.5cm,height=8cm]{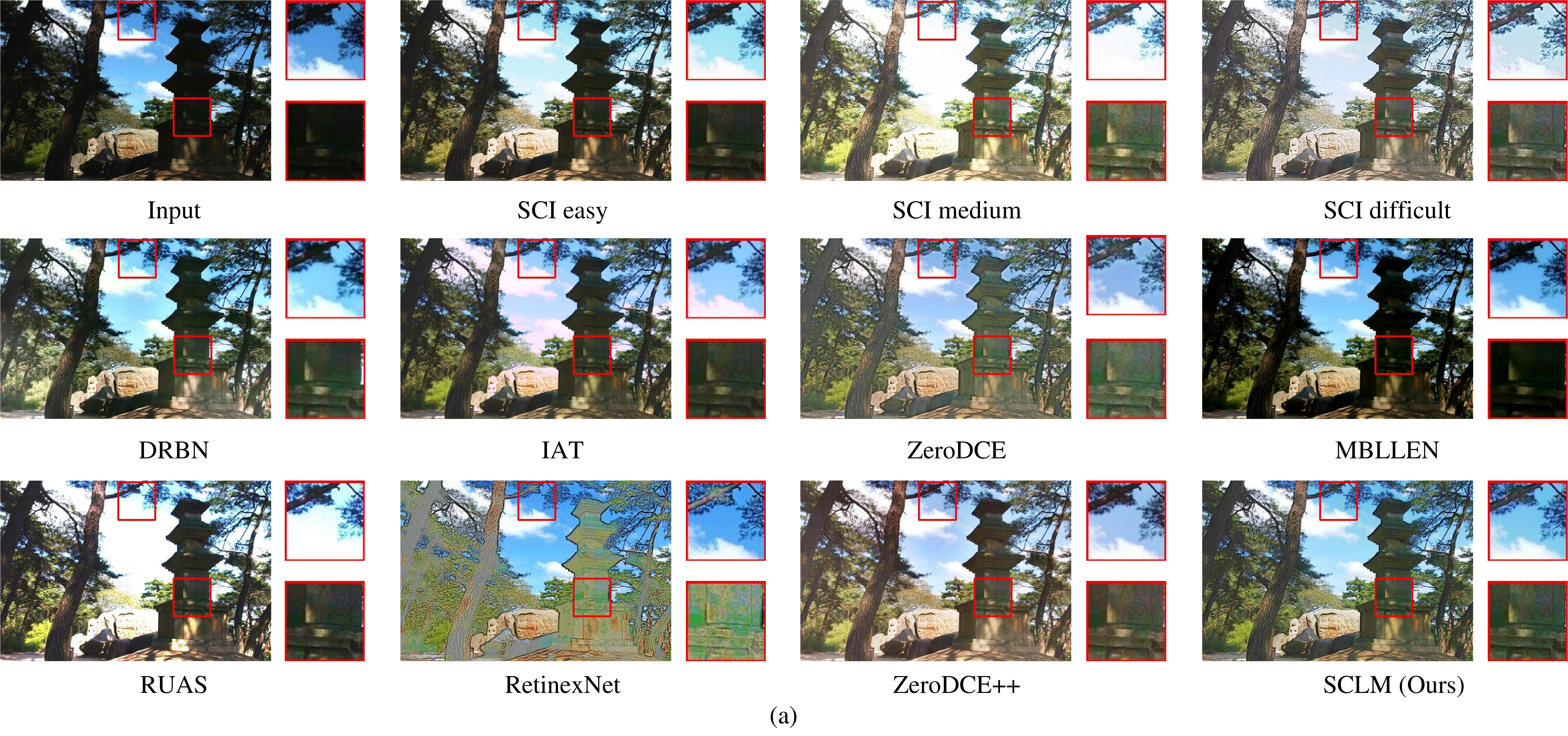}
\end{figure*}
\begin{figure*}[htbp]
		\vspace{-2em}
	\centering
	\includegraphics[width=16.5cm,height=8cm]{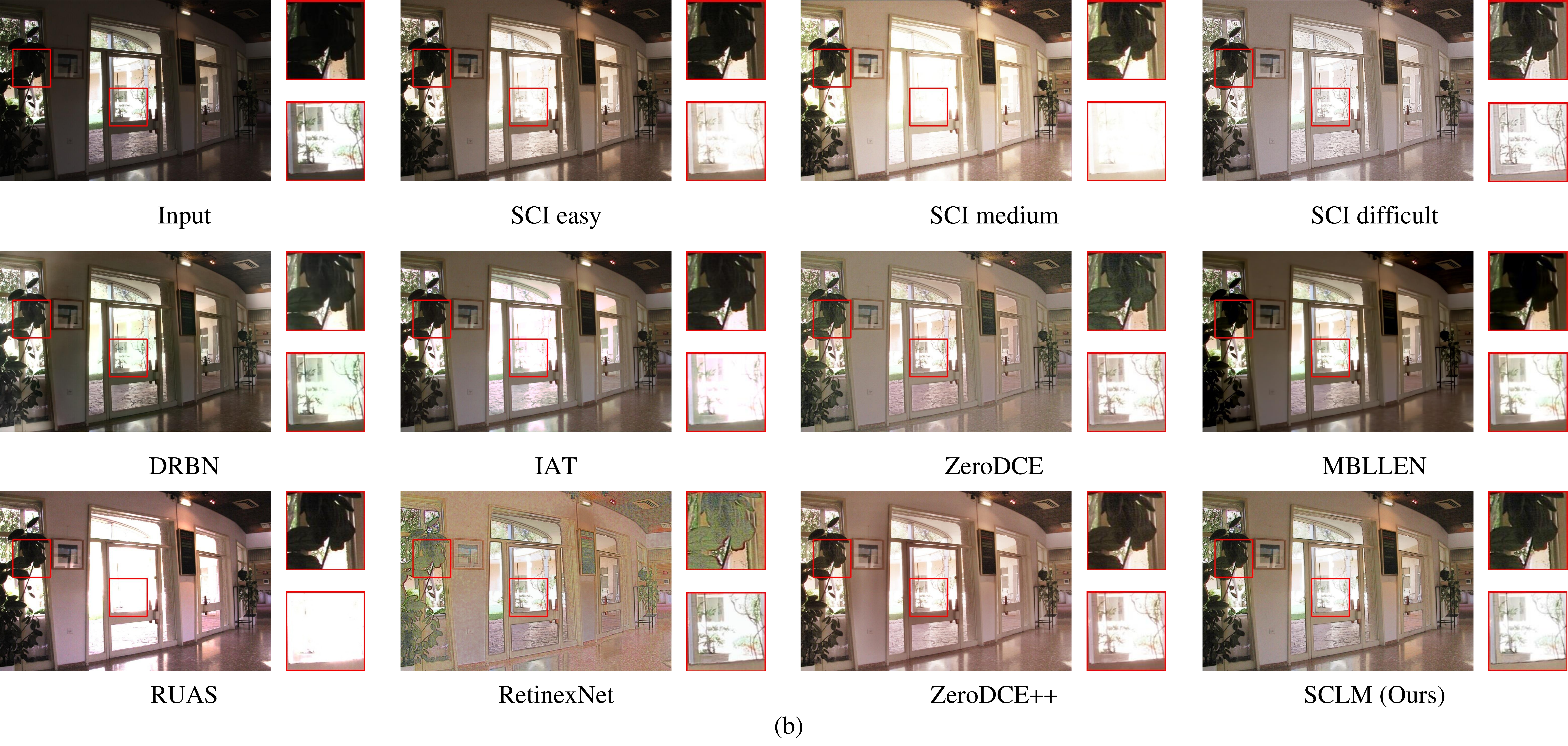}
	\caption{ Visual comparison of different methods on DICM\cite{DBLP:conf/icip/LeeLK12}. Parts of areas are zoomed in with red boxes.} \label{fig:DICM_2}
\end{figure*}
\subsection{Implementation Details}
The proposed model is trained on the multi-exposure dataset\cite{DBLP:conf/cvpr/AfifiDOB21}. %
This dataset is captured with the Adobe Camera
Raw SDK\cite{Adobe} to emulate different digital exposure values (EVs)  applied
by a camera. They apply varying EVs to each raw-RGB image to simulate actual exposure errors. This involves using relative EVs of -1.5, -1, 0, +1, and +1.5 to produce images with underexposure errors, no gain from the original EV, and overexposure errors, respectively. As the ground truth images, they manually retouched 0 EVs images by an expert photographer.
Since our aim is the low-light image enhancement, we only use data with -1.5EV, -1 EV, and 0 Ev as input and do not use overexposed images. The evaluations are conducted on five public real-world shot datasets to simulate real-world low-light scenarios better.    Specifically, following the ZeroDCE series, we perform experiments on NPE\cite{Naturalness2013}, LIME\cite{LIME2017}, MEF\cite{DBLP:journals/tip/MaZW15}, DICM
\cite{DBLP:conf/icip/LeeLK12}, VV\cite{vonikakis2021busting}. We also calculate the average illuminance (Y channel) of these five datasets in Table~\ref{tab:2}, It can be seen that there is considerable variation in the average illuminance of the five datasets. Datasets with multiple lighting conditions allow for better testing of the generalization capabilities of the different methods.   Since low-light images in real scenes often do not have corresponding ground truth, we employ two no-reference indicators, PI\cite{DBLP:conf/cvpr/BlauM18,DBLP:journals/cviu/MaYY017,DBLP:journals/spl/MittalSB13} and NIQE\cite{DBLP:journals/spl/MittalSB13} as evaluation metrics, which are also adopted by many recent LLIE methods\cite{SCI2022,2020Zero,2022Zero}.

 We implement our models with PyTorch on an
NVIDIA 1080 GPU. 
During training, we crop the images into 256$\times$256 patches and set the batch size as 16. Data augmentation operations such as horizontal/vertical flipping and rotation are also performed. The training process is carried out in a fully supervised manner, i.e., the model takes in low-light images as input and is supervised by corresponding ground truth images.  $\mathcal{L}_{1}$ loss is adopted during the training process. For the curve ajustment parameters, we initialize $\alpha$, $\beta$, and $\gamma$ as 0.6, -1.3, 1.5 respectively. 
This means that the quadratic function is initialized to: $C = 0.6*x^{2}-1.3*x + 1.5 $. Under this setting, we have: $C(0) = 1.5$, $C(0.5) = 1$,$C(1) = 0.8$. The motivation for setting the initial value in this way is that we hope the areas with relatively low lightness can be further enhanced and, meanwhile, suppress overexposure in relatively brighter areas. 
  \begin{table}[t]
	\caption{ Average illuminance on Y channel of different datasets. }
	\centering
	\resizebox{0.95\linewidth}{!}{
		\begin{tabular}{c|ccccc}
			\hline
			Dataset              & LIME  & VV    & MEF   & NPE   & DICM \\ \hline
			Average illumination  & 38.61 & 66.29 & 41.06 & 87.04 & 84.16  \\ \hline
		\end{tabular}\label{tab:2}
	}
\end{table}
We utilize the Adam \cite{kingma2014adam} with $\beta_{1}$ = 0.9 and $\beta_{2}$ = 0.999 as optimizer. The learning rate is initialized as  
2 $\times$ 10$^{-4}$ and is decayed to 1 $\times$ 10$^{-6}$ with a cosine annealing scheduler\cite{DBLP:conf/iclr/LoshchilovH17}.
Additionally, our model's lightweight design facilitates a highly efficient training process, resulting in significant time savings and reduce environmental impact with a low carbon footprint. Unlike those huge models that require dozens of hours for training on multiple GPUs, our network converges in less than an hour with just one NVIDIA 1080 card.

\subsection{Benchmark Evaluations} 
We conduct comparisons with state-of-the-art
methods, including traditional method (LIME\cite{LIME2017}), four
supervised learning based methods (RetinexNet\cite{wei2018deep}, MBLLEN\cite{DBLP:conf/bmvc/LvLWL18}, DRBN\cite{DBLP:conf/cvpr/Yang0FW020}, IAT\cite{DBLP:conf/bmvc/CuiL0SG00H22}) and four zero-shot learning method (SCI\cite{SCI2022}, RUAS\cite{RUS2021}, ZeroDCE\cite{2020Zero}, ZeroDCE++\cite{2022Zero}). We utilize publicly available source codes and recommended parameters to reproduce the results.

 The quantitative comparisons are listed in Table~\ref{tab:1}. As one can see, SCLM achieves the best results in terms of average PI  and obtain the second-best performance in terms of average NIQE on five datasets. Meanwhile, SCLM outperforms all the other competitors on all indicators of NPE and DICM. These two datasets contain more diverse and complex lighting scenarios (i.e., containing both dark and light areas). The measurement results show that our method is quite competitive, especially when dealing with complex lighting conditions.

From  Fig.~\ref{fig:LIME_2} to Fig.~\ref{fig:DICM_2}, we compare the enhancement results for five real-world datasets. 
We provide two sets of enhanced results for each dataset.  Subjective results  include a variety of situations, including indoor and outdoor scenes, as well as images with different exposure levels. We will first give a few specific analyses of typical scenes and then give a summary.
We start by analyzing the enhancement results for images taken in poor daytime lighting conditions, where the shooting angle often causes parts of the image to be too dark.
In VV (a), we display a set of scenes containing portraits taken under low-light conditions. The SCI series and RUAS exhibit uneven exposure. The Retinexnet and IAT appear to have color deviation. MBLLEN and DRBN tend to under-enhance the low-light input. In comparison, our method and ZeroDCE perform relatively well. Similar to VV (a), in MEF (a), we show a landscape photo taken in backlit conditions. The SCI series and RUAS exhibit uneven exposures. IAT, RetinexNet, and the ZeroDCE series appear to have color deviation.
Similar situations can also be observed in NPE (a), NPE (b), and DICM (a).
Another typical situation is the dark scene containing locally bright areas, such as an image taken at night with point lights. We observe that some methods may produce significant overexposure of the bright regions. For example, in LIME (a), the enhancement results of RUAS and SCI medium produce overexposure in some bright light areas. Another  example is given in MEF (b), where methods including SCI medium, DRBN, IAT, and RUAS all result in overexposure of the point light source, with the overexposed area covering the lampshade. In contrast, curve-based methods can suppress overexposure to some extent, but some of them also result in color deviation. (For example, the enhancement results of the ZeroDCE series in MEF (b).) Thanks to the local adaptation strategy, our method can cope better with these scenes.

  \begin{table}[t]
 	\caption{Perceptual index (PI) $\downarrow$ / naturalness image quality evaluator (NIQE)  $\downarrow$ of w/wo local adaptation module(LAM). }\label{tab:3}
 	\centering
 	\resizebox{1.01\linewidth}{!}{
 		\begin{tabular}{c|ccccll}
 			\hline
 			& LIME      & VV        & MEF       & NPE       & \multicolumn{1}{c}{DICM} & \multicolumn{1}{c}{Average} \\ \hline
 			w/o LAM & 3.08/4.25 & 3.46/3.27 & 3.07/3.86 & 2.78/3.76 & 3.47/4.08                & 3.17/3.84                   \\ \hline
 			w/ LAM  & 2.85/4.06 & 2.86/2.65 & 2.58/3.57 & 2.75/3.77 & 2.83/3.69                & 2.77/3.55                   \\ \hline
 		\end{tabular}\label{tab:4}
 	}
 \end{table}
 Although all methods, including our SCLM, cannot handle all scenarios perfectly, our approach is very competitive regarding overall results, especially considering its extremely low complexity. Some methods produce unpleasant artifacts when working with complex low-light scenes, including uneven exposure, color deviation, and color undersaturation. Specifically, earlier methods such as RetinexNet, and DRBN tend to produce unnatural artifacts. The SCI series and RUAS are less robust and tend to produce unevenly exposed results. MBLLEN, on the other hand, shows excessive smoothing and blurring in some natural scenes. Both our method and ZeroDCE produce
 relatively color-coordinated results, but ZeroDCE may sometimes produce color bias. Meanwhile, ZeroDCE requires calculating per-pixel adjustment parameters and needs to be iteratively correct, which is particularly unfriendly for high-resolution images, while our method shares a set of adjustment parameters globally.

  \begin{figure*}[htbp]
  		\vspace{-1em}
 	\centering
 	\includegraphics[width=17.5cm]{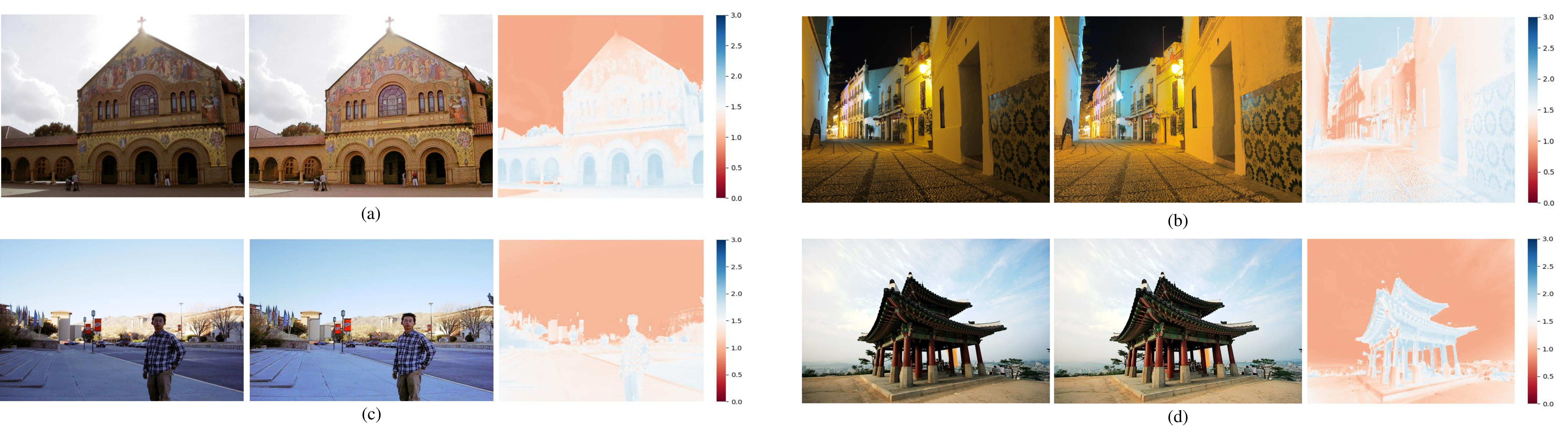}
 	\caption{Visual comparison of w/wo local adaptation module. We  also display the heatmap of the local adaptation map. It can be seen that the details in some regions of the image have been further enhanced after local adjustment.} \label{fig:6}
 \end{figure*}
 
\subsection{Ablation study} 

\noindent \textbf{Local Adaptation module.} 
To validate the effectiveness of the local adaptive module, we remove this part and train two models with the same training configuration. Results are compared on five real-world datasets as mentioned above. As shown in Table~\ref{tab:3}, the results after local adjustment are better than those without local adjustment in most of the evaluation metrics. Subjective visual results also confirm this. We show some typical examples in Fig.~\ref{fig:6}.  

For each set of images, we show from left to right the enhancement results  of without and with the local adaptation module and the corresponding adaptation map. It can be seen that the enhanced outputs with the inclusion of local adaptation map
($C$ in Equ.~\ref{f6}) have better subjective results, especially for some backlit areas, such as the backlit face of the church in (a), and the person's facial details in (c).
   \begin{figure*}[t]
	\centering
	\includegraphics[width=17.5cm]{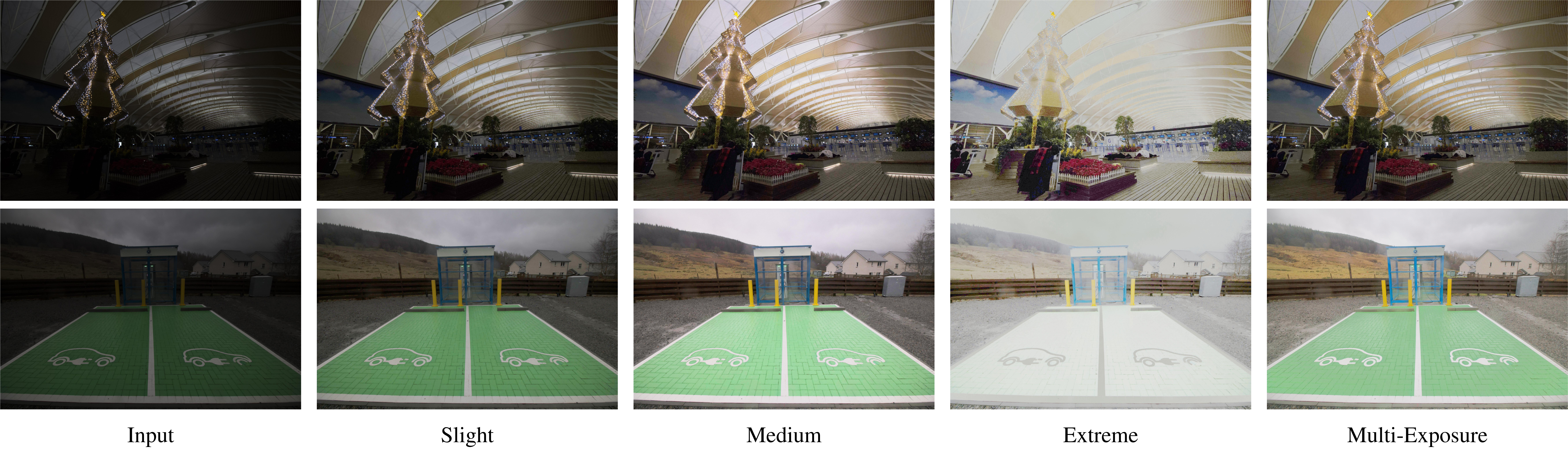}
	\caption{
		Visual comparison on the impact of training data. Training with muti-exposure dataset  can lead to better restoration results.
	} \label{fig:8}
\end{figure*}
 \begin{figure}[t]
 		\centering
 		\includegraphics[width=8.5cm]{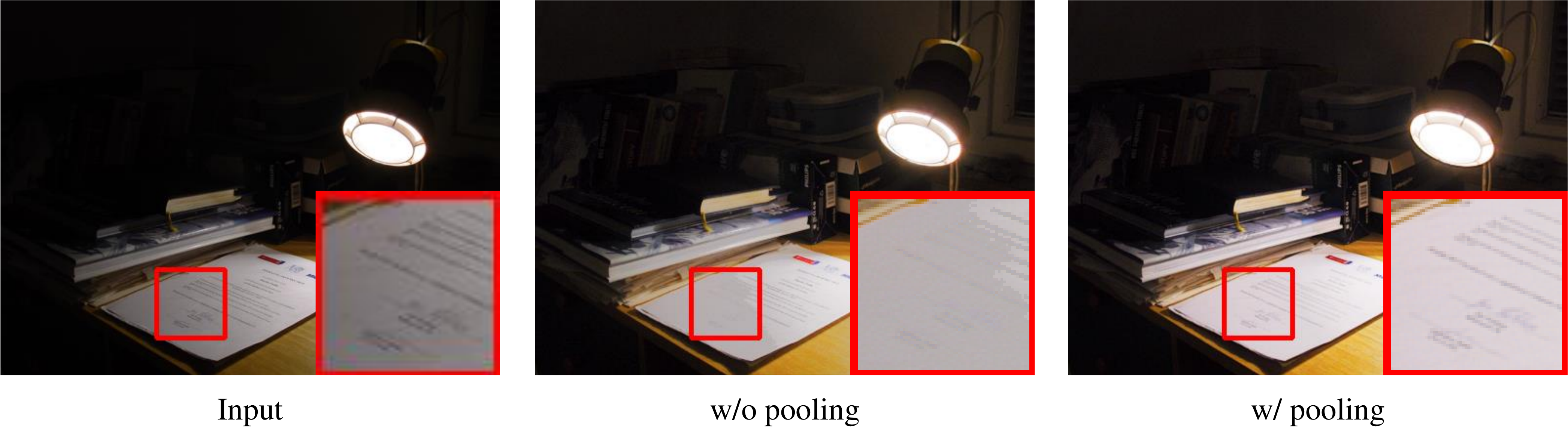}
 		\caption{Visual comparison of w/wo max pooling operation. } \label{fig:7}
 \end{figure}
\begin{table}[t]
	\caption{ Perceptual index (PI) $\downarrow$ / naturalness image quality evaluator (NIQE)  $\downarrow$ of different structural re-parameterization topology. AB,TD and DB refers to Asymmetric Block, Triple Duplicate and Diverse Branch respectively.}\label{tab:4}
	\centering
	\resizebox{1.01\linewidth}{!}{
		\begin{tabular}{c|cccccc}
			\hline
			& LIME      & VV        & MEF       & NPE       & DICM      & Average   \\ \hline
			AB & 2.95/4.33 & 3.10/2.86 & 2.73/3.86 & 2.97/4.18 & 2.95/3.89 & 2.94/3.83 \\
			TD & 2.88/4.17 & 3.01/2.74 & 2.59/3.66 & 2.67/3.65 & 2.83/3.64 & 2.80/3.57 \\
			DB             & 2.85/4.06 & 2.86/2.65 & 2.58/3.57 & 2.75/3.77 & 2.83/3.69 & 2.77/3.55 \\ \hline
		\end{tabular}
		
	}
\end{table}
 We emphasize that it is significantly better to incorporate the local adjustment module for the coarsely enhanced image. This is because the receptive field of the Retinex-based CNN model is very limited, e.g., for a 3$\times$3 convolution, it can only cover three adjacent pixels and lacks global brightness prior. Some work \cite{DBLP:conf/bmvc/CuiL0SG00H22} has introduced transformer architecture to increase the receptive field, but the computational cost of this method is very high. In contrast, we introduce the relative brightness prior to the local area in a straightforward but effective way, enabling the model to obtain satisfactory enhancement results quickly and efficiently.
 Next, we explain the effect of performing maximum pooling on the adaptation map. This operator can alleviate excessive enhancement for dark-colored pixels, especially for some black text parts. A typical example is given in Fig.~\ref{fig:7}, where it can be seen that the overexposure of the text part is significantly relieved after pooling. It should be noted that this operation is actually only effective for some scenarios. How to prevent excessive enhancement of dark-colored objects (such as black hair) leading to color deviation is still an open problem in the low-light enhancement field. Adding semantic information may alleviate this phenomenon, but we focus on ultra-lightweight low-light enhancement networks here, so we do not consider it.
 To summarize, the local adaptation module brings considerable visual improvement with only three learnable parameter. This might suggest that researchers should  consider more image brightness priors for low-light enhancement tasks rather than designing complicated structures. \\
 \noindent \textbf{Structure topology.}

  \begin{figure}[t]
 	\centering
 	\includegraphics[width=9.0cm]{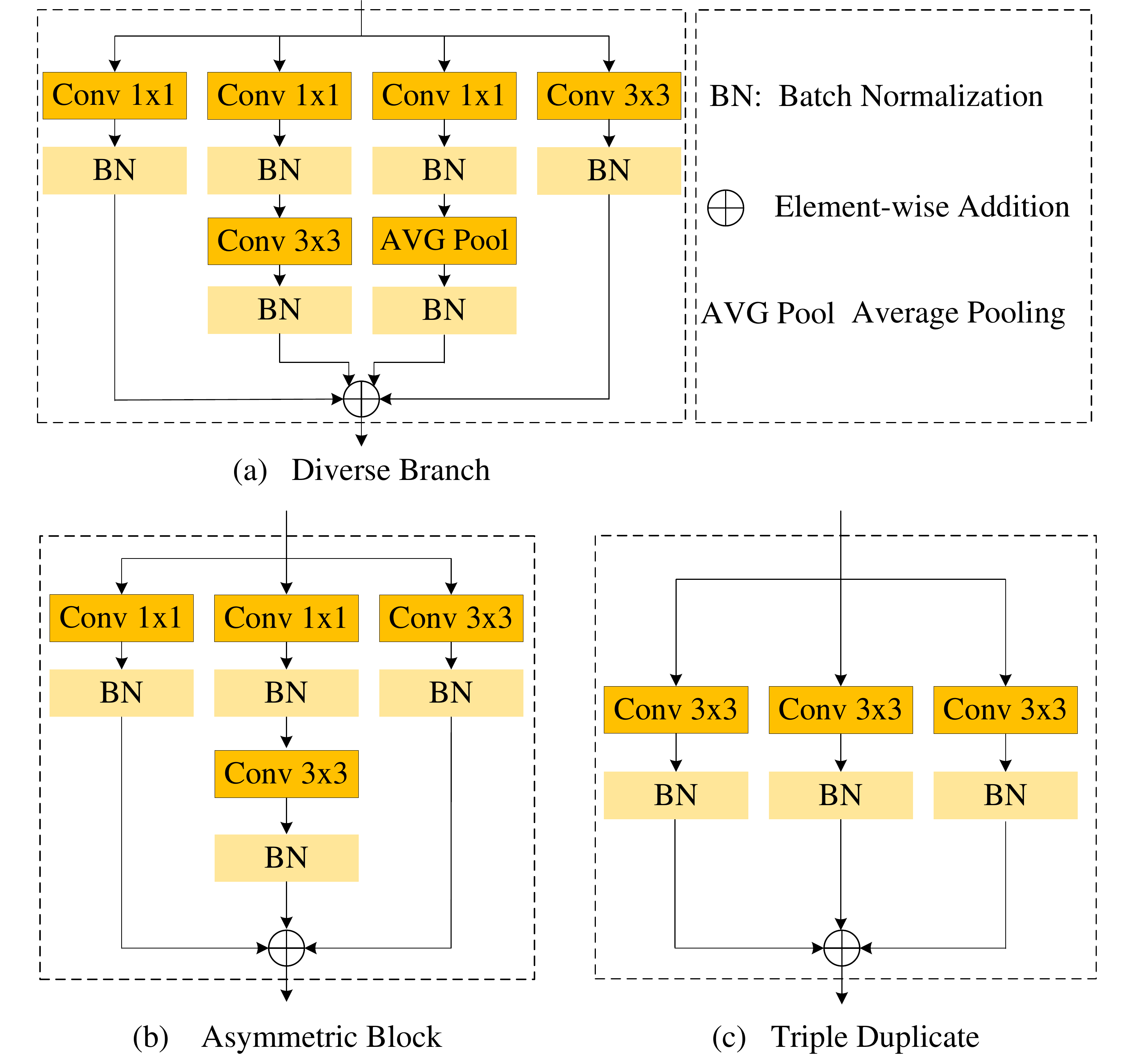}
 	\caption{Comparison of different multi-branch topologies. We name these three structures Diverse Branch, Asymmetric Block, and Triple Duplicate. The parts in the dashed box can all be equivalently converted to a 3$\times$3 convolution.
 	} \label{fig:9}
 \end{figure}
 To determine which structure is more suitable for estimating illuminance component, we test several multi-branch structures. As shown in Fig.~\ref{fig:8}, we dub these three structures  Diverse Branch (DB), Asymmetric Block (AB), and Triple Duplicate (TD), respectively. The part in the dashed box can be collapsed into a 3$\times$3 convolution equivalently. We train and test these three structures on the muti-exposure dataset\cite{DBLP:conf/cvpr/AfifiDOB21}, and the test results are shown in Table~\ref{tab:4}. It can be seen that the Diverse Branch (DB) can achieve the best performance, so we choose DB as the super-structure for training. In fact, we find that these multi-branch structures can achieve similar perceptual results except for the plain structure, which has a significantly  weaker modeling ability.
 
\subsection{Impact of training data} 
To test the impact of training
data, we retrain our model on SICE dataset part1\cite{DBLP:journals/tip/CaiGZ18} with different settings. 
SICE contains sets of images with different exposure levels, and we only adopt the underexposed images and their corresponding well-lit counterpart as training data. To be specific, we select three images with different brightness levels for each image series and thus obtain three sub-datasets, which we dub slight-low-light, medium-low-light, and extreme-low-light. Through experiment, We find that the training data's luminance greatly affects the low light enhancement results. Training with the three datasets resulted in three models with different low-light enhancement capabilities. We compare the enhanced results of these three models with the results of the model trained with multiple exposure datasets\cite{DBLP:conf/cvpr/AfifiDOB21}.
As shown in Fig.~\ref{fig:8}, the enhancement results of  $model_{slight}$ have relatively low luminance, while $model_{extreme}$ tends to over-enhance the overall image.  And the enhanced brightness of the $model_{medium}$ is somewhere between $model_{slight}$  and $model_{extreme}$. In contrast, the results of training with multiple exposure data produce better visual results. The results demonstrate that incorporating multi-exposure training data in the training stage is very necessary.

 \begin{table}[t]
	\caption{ Computation efficiency of different methods.  The inference time and GFLOPS are measured when processing 1080P image. The best results are highlighted with {\color{red} red} color, and the second best results are  highlighted with {\color{blue} blue} color }
	\centering
	\resizebox{0.96\linewidth}{!}{

		\begin{tabular}{c|cccc}
			\hline
			Methods    & Parameter(M) & GFLOPS & GPU Time & \multicolumn{1}{l}{CPU Time} \\ \hline
			LIME\cite{LIME2017}       & -         & -      & -        & 255.35                       \\
			RetinexNet\cite{wei2018deep} & 0.555     & 640.05 & 0.90     & 10.29                        \\
			MBLLEN \cite{DBLP:conf/bmvc/LvLWL18}    & 0.450     & 930.92 & 0.32     & 33.73                        \\
			DRBN\cite{DBLP:conf/cvpr/Yang0FW020}       & 0.557     & 335.49 & 1.57     & 10.98                        \\
			IAT\cite{DBLP:conf/bmvc/CuiL0SG00H22}         & 0.087     & 42.29  & 0.64     & 10.11                        \\
			RUAS\cite{RUS2021}       & 0.003     & 6.77   & 0.11     & 2.50                         \\
			SCI\cite{SCI2022}         & \textcolor{blue}{0.0003}    & 1.11   & \textcolor{red}{0.01}     & \textcolor{blue}{0.13}                         \\
			ZeroDCE\cite{2020Zero}    & 0.079     & 164.23 & 0.12     & 2.48                         \\
			ZeroDCE++\cite{2022Zero}  & 0.011     & \textcolor{blue}{0.21}   & \textcolor{red}{0.01}     & 0.28                         \\
			SCLM (Ours)       & \textcolor{red}{0.000087}  &\textcolor{red}{ 0.17}   & \textcolor{blue}{0.02}     & \textcolor{red}{0.12}                         \\ \hline
		\end{tabular}
	}\label{tab:5}
\end{table}

\subsection{Speed and efficiency}

Next, we compare the proposed method's inference speed and complexity with other approaches. Specifically, we compare the efficiency of different methods in processing 1080P images (1080$\times$1920), and the results are shown in Table~\ref{tab:5}.

Considering the different devices available under different conditions, we test both CPU and GPU inference times. The CPU used for the tests is an Intel(R) Xeon(R) Silver 4310 CPU with 2.10GHz. And the GPU is an NVIDIA 1080 card. We first compare and analyze the model size. As shown in Table~\ref{tab:5}, our model has the minimum number of parameters. It only contains one convolutional layer and three modulation coefficients, totaling 87 parameters. Some other methods (RUAS\cite{RUS2021}  and SCI\cite{SCI2022}) are also very lightweight. This suggests that large models with hundred layers may not be necessary for many low-light sceneries. At the same time, our method also has the least  Giga Floating-Point Operations Per Second (GFLOPS) when processing images, indicating its great advantage in power consumption. However, it's worth noting that our method is not the fastest method on GPU. It slightly suffers when compared with SCI and ZeroDCE++. For SCI, it only contains three 3$\times$3 Conv layers which 
are highly optimized by modern deep learning frameworks. For ZeroDCE++, it downsamples the input image when estimating the high-order curve. In contrast, our local adaptation module is relatively complex and has not been specifically optimized. However, both subjective and objective results verify that these two methods perform worse than our approach. The slight additional time in exchange for better visual quality is thus justifiable.  When running on the CPU, the inference speed of all methods is significantly reduced, and our method achieves the fastest inference speed. These results show that our method is well adapted to different devices and achieves the best trade-off between speed and effectiveness. In fact, similar to ZeroDCE++, our approach still has optimization space. For example, when processing high-resolution images, we can downsample the image first, calculate the corresponding illumination component and local adaptation map for the downscaled image, and balance speed and effect. Even without any optimization, our approach can handle low-light input images in real time on an ordinary graphics card(e.g., NVIDIA 1080).  Another interesting finding is that some traditional methods (such as LIME\cite{LIME2017}) have decent performance but are very slow due to the lack of corresponding parallel optimization and iterative calculations. This once again illustrates the huge potential of combining traditional methods and deep learning strategies for  light adaptation tasks.
   \begin{figure}[t]
	\centering
	\includegraphics[width=9.0cm]{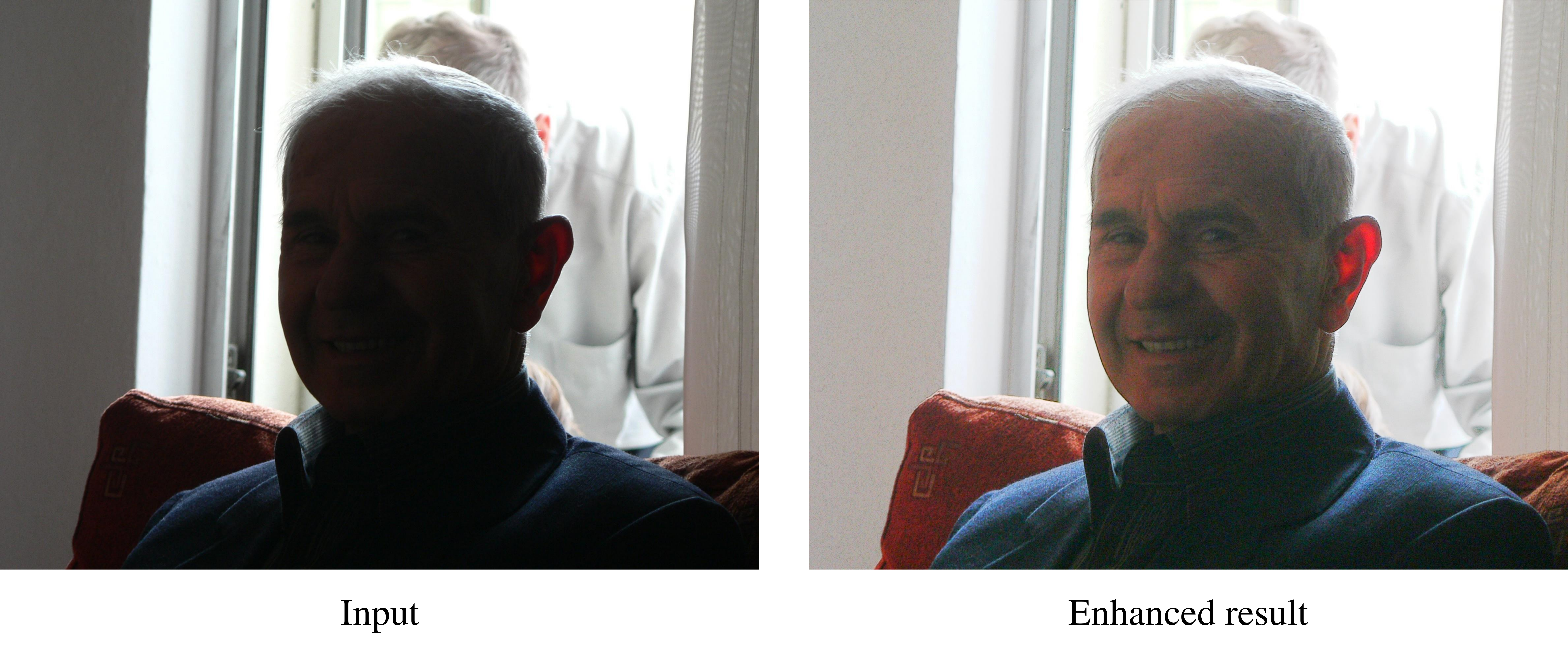}
	\caption{Limitation of the proposed method. our model cannot cope with the images with serious noise well (e.g., The face region in the enhanced image appears obvious noise). Zoom in  for a better visualization.
	} \label{fig:10}
\end{figure}
\subsection{Limitations}
Although the proposed one-layer model has achieved promising results, some limitations still exist.
The first limitation is that our model often fails to achieve satisfactory results when processing scenes with a lot of noise. We display a typical example in Fig.~\ref{fig:10}. This is because a one-layer model cannot compete with deeper models that have hundreds of Conv layers with stronger denoising capabilities. Another limitation is that although we have taken measures to avoid unnatural enhancement in local regions of images, there are still very few cases where a picture contains both very bright and very dark regions, resulting in unnatural color and color deviation in the enhanced result. This problem  remains unsolved with most of the current methods. We leave these problems for our future work.

\section{Conclusions}
 This paper presents a lightweight structure for low-light image enhancement which contains only a single convolutional layer.
We first explore the complexity extreme of learning-based LLIE methods. With the incorporation of structural re-parameterization techniques, we find that a single convolutional layer can achieve promising enhancement results. Then, inspired by curve adjustment schemes, a local adaptation module is further introduced to better adjust the brightness of local regions. The parameter-sharing strategy and the one-adjustment-only scheme also ensure its efficiency. Experimental results show that the proposed method is comparable with state-of-the-art methods in real-world low-light image enhancement tasks with fewer parameters and lower computational complexity.

\bibliographystyle{IEEEtran}
\bibliography{IEEEabrv,./bibliography}

\begin{thebibliography}{10}
\providecommand{\url}[1]{#1}
\csname url@samestyle\endcsname
\providecommand{\newblock}{\relax}
\providecommand{\bibinfo}[2]{#2}
\providecommand{\BIBentrySTDinterwordspacing}{\spaceskip=0pt\relax}
\providecommand{\BIBentryALTinterwordstretchfactor}{4}
\providecommand{\BIBentryALTinterwordspacing}{\spaceskip=\fontdimen2\font plus
\BIBentryALTinterwordstretchfactor\fontdimen3\font minus
  \fontdimen4\font\relax}
\providecommand{\BIBforeignlanguage}[2]{{%
\expandafter\ifx\csname l@#1\endcsname\relax
\typeout{** WARNING: IEEEtran.bst: No hyphenation pattern has been}%
\typeout{** loaded for the language `#1'. Using the pattern for}%
\typeout{** the default language instead.}%
\else
\language=\csname l@#1\endcsname
\fi
#2}}
\providecommand{\BIBdecl}{\relax}
\BIBdecl

\bibitem{stark2000adaptive}
J.~A. Stark, ``Adaptive image contrast enhancement using generalizations of
  histogram equalization,'' \emph{{IEEE} Trans. Image Process.}, vol.~9, no.~5,
  pp. 889--896, 2000.

\bibitem{coltuc2006exact}
D.~Coltuc, P.~Bolon, and J.~Chassery, ``Exact histogram specification,''
  \emph{{IEEE} Trans. Image Process.}, vol.~15, no.~5, pp. 1143--1152, 2006.

\bibitem{ibrahim2007brightness}
H.~Ibrahim and N.~S.~P. Kong, ``Brightness preserving dynamic histogram
  equalization for image contrast enhancement,'' \emph{{IEEE} Trans. Consumer
  Electron.}, vol.~53, no.~4, pp. 1752--1758, 2007.

\bibitem{rahman2004retinex}
Z.-u. Rahman, D.~J. Jobson, and G.~A. Woodell, ``Retinex processing for
  automatic image enhancement,'' \emph{Journal of Electronic imaging}, vol.~13,
  no.~1, pp. 100--110, 2004.

\bibitem{Naturalness2013}
S.~Wang, J.~Zheng, H.~Hu, and B.~Li, ``Naturalness preserved enhancement
  algorithm for non-uniform illumination images,'' \emph{{IEEE} Trans. Image
  Process.}, vol.~22, no.~9, pp. 3538--3548, 2013.

\bibitem{Probabilistic2015}
X.~Fu, Y.~Liao, D.~Zeng, Y.~Huang, X.~S. Zhang, and X.~Ding, ``A probabilistic
  method for image enhancement with simultaneous illumination and reflectance
  estimation,'' \emph{{IEEE} Trans. Image Process.}, vol.~24, no.~12, pp.
  4965--4977, 2015.

\bibitem{wei2018deep}
C.~Wei, W.~Wang, W.~Yang, and J.~Liu, ``Deep retinex decomposition for
  low-light enhancement,'' in \emph{British Machine Vision Conference,
  {BMVC}}.\hskip 1em plus 0.5em minus 0.4em\relax {BMVA} Press, 2018, p. 155.

\bibitem{DBLP:conf/cvpr/Yang0FW020}
W.~Yang, S.~Wang, Y.~Fang, Y.~Wang, and J.~Liu, ``From fidelity to perceptual
  quality: {A} semi-supervised approach for low-light image enhancement,'' in
  \emph{Conference on Computer Vision and Pattern Recognition, {CVPR}}.\hskip
  1em plus 0.5em minus 0.4em\relax Computer Vision Foundation / {IEEE}, 2020,
  pp. 3060--3069.

\bibitem{DBLP:journals/tip/JiangGLCFSYZW21}
Y.~Jiang, X.~Gong, D.~Liu, Y.~Cheng, C.~Fang, X.~Shen, J.~Yang, P.~Zhou, and
  Z.~Wang, ``Enlighten{GAN}: Deep light enhancement without paired
  supervision,'' \emph{{IEEE} Trans. Image Process.}, vol.~30, pp. 2340--2349,
  2021.

\bibitem{DBLP:conf/aaai/WangWYLCK22}
Y.~Wang, R.~Wan, W.~Yang, H.~Li, L.~Chau, and A.~C. Kot, ``Low-light image
  enhancement with normalizing flow,'' in \emph{Thirty-Sixth {AAAI} Conference
  on Artificial Intelligence, {AAAI}}.\hskip 1em plus 0.5em minus 0.4em\relax
  {AAAI} Press, 2022, pp. 2604--2612.

\bibitem{RUS2021}
R.~Liu, L.~Ma, J.~Zhang, X.~Fan, and Z.~Luo, ``Retinex-inspired unrolling with
  cooperative prior architecture search for low-light image enhancement,'' in
  \emph{Conference on Computer Vision and Pattern Recognition, {CVPR}}.\hskip
  1em plus 0.5em minus 0.4em\relax {Computer Vision Foundation / {IEEE}}, 2021,
  pp. 10\,561--10\,570.

\bibitem{SCI2022}
L.~Ma, T.~Ma, R.~Liu, X.~Fan, and Z.~Luo, ``Toward fast, flexible, and robust
  low-light image enhancement,'' in \emph{Conference on Computer Vision and
  Pattern Recognition, {CVPR}}.\hskip 1em plus 0.5em minus 0.4em\relax Computer
  Vision Foundation / {IEEE}, 2022, pp. 5627--5636.

\bibitem{2020Zero}
C.~Guo, C.~Li, J.~Guo, C.~C. Loy, J.~Hou, S.~Kwong, and R.~Cong,
  ``Zero-reference deep curve estimation for low-light image enhancement,'' in
  \emph{Conference on Computer Vision and Pattern Recognition, {CVPR}}.\hskip
  1em plus 0.5em minus 0.4em\relax Computer Vision Foundation / {IEEE}, 2020,
  pp. 1777--1786.

\bibitem{2022Zero}
C.~Li, C.~Guo, and C.~C. Loy, ``Learning to enhance low-light image via
  zero-reference deep curve estimation,'' \emph{{IEEE} Trans. Pattern Anal.
  Mach. Intell.}, vol.~44, no.~8, pp. 4225--4238, 2022.

\bibitem{DBLP:conf/cvpr/BlauM18}
Y.~Blau and T.~Michaeli, ``The perception-distortion tradeoff,'' in
  \emph{Conference on Computer Vision and Pattern Recognition, {CVPR}}.\hskip
  1em plus 0.5em minus 0.4em\relax Computer Vision Foundation / {IEEE}, 2018,
  pp. 6228--6237.

\bibitem{DBLP:journals/cviu/MaYY017}
C.~Ma, C.~Yang, X.~Yang, and M.~Yang, ``Learning a no-reference quality metric
  for single-image super-resolution,'' \emph{Comput. Vis. Image Underst.}, vol.
  158, pp. 1--16, 2017.

\bibitem{DBLP:journals/spl/MittalSB13}
A.~Mittal, R.~Soundararajan, and A.~C. Bovik, ``Making a "completely blind"
  image quality analyzer,'' \emph{{IEEE} Signal Process. Lett.}, vol.~20,
  no.~3, pp. 209--212, 2013.

\bibitem{land1977retinex}
E.~H. Land, ``The retinex theory of color vision,'' \emph{Scientific american},
  vol. 237, no.~6, pp. 108--129, 1977.

\bibitem{LIME2017}
X.~Guo, Y.~Li, and H.~Ling, ``{LIME:} low-light image enhancement via
  illumination map estimation,'' \emph{{IEEE} Trans. Image Process.}, vol.~26,
  no.~2, pp. 982--993, 2017.

\bibitem{lee2013contrast}
C.~Lee, C.~Lee, and C.~Kim, ``Contrast enhancement based on layered difference
  representation of 2d histograms,'' \emph{{IEEE} Trans. Image Process.},
  vol.~22, no.~12, pp. 5372--5384, 2013.

\bibitem{survey2022}
C.~Li, C.~Guo, L.~Han, J.~Jiang, M.~Cheng, J.~Gu, and C.~C. Loy, ``Low-light
  image and video enhancement using deep learning: {A} survey,'' \emph{{IEEE}
  Trans. Pattern Anal. Mach. Intell.}, vol.~44, no.~12, pp. 9396--9416, 2022.

\bibitem{shen2017msr}
L.~Shen, Z.~Yue, F.~Feng, Q.~Chen, S.~Liu, and J.~Ma, ``Msr-net: Low-light
  image enhancement using deep convolutional network,'' \emph{arXiv preprint
  arXiv:1711.02488}, 2017.

\bibitem{DBLP:journals/tcsv/ZhaoXWOYK22}
Z.~Zhao, B.~Xiong, L.~Wang, Q.~Ou, L.~Yu, and F.~Kuang, ``Retinex{DIP}: {A}
  unified deep framework for low-light image enhancement,'' \emph{{IEEE} Trans.
  Circuits Syst. Video Technol.}, vol.~32, no.~3, pp. 1076--1088, 2022.

\bibitem{DBLP:journals/tcsv/FanFGCC22}
G.~Fan, B.~Fan, M.~Gan, G.~Chen, and C.~L.~P. Chen, ``Multiscale low-light
  image enhancement network with illumination constraint,'' \emph{{IEEE} Trans.
  Circuits Syst. Video Technol.}, vol.~32, no.~11, pp. 7403--7417, 2022.

\bibitem{DBLP:journals/tcsv/LiFH21}
J.~Li, X.~Feng, and Z.~Hua, ``Low-light image enhancement via
  progressive-recursive network,'' \emph{{IEEE} Trans. Circuits Syst. Video
  Technol.}, vol.~31, no.~11, pp. 4227--4240, 2021.

\bibitem{DBLP:journals/tcsv/LiuWW22}
C.~Liu, F.~Wu, and X.~Wang, ``Efinet: Restoration for low-light images via
  enhancement-fusion iterative network,'' \emph{{IEEE} Trans. Circuits Syst.
  Video Technol.}, vol.~32, no.~12, pp. 8486--8499, 2022.

\bibitem{DBLP:conf/cvpr/AfifiDOB21}
M.~Afifi, K.~G. Derpanis, B.~Ommer, and M.~S. Brown, ``Learning multi-scale
  photo exposure correction,'' in \emph{Conference on Computer Vision and
  Pattern Recognition, {CVPR}}.\hskip 1em plus 0.5em minus 0.4em\relax Computer
  Vision Foundation / {IEEE}, 2021, pp. 9157--9167.

\bibitem{DBLP:conf/iccv/DingGDH19}
X.~Ding, Y.~Guo, G.~Ding, and J.~Han, ``{ACN}et: Strengthening the kernel
  skeletons for powerful {CNN} via asymmetric convolution blocks,'' in
  \emph{2019 {IEEE/CVF} International Conference on Computer Vision,
  {ICCV}}.\hskip 1em plus 0.5em minus 0.4em\relax {IEEE}, 2019, pp. 1911--1920.

\bibitem{DBLP:conf/cvpr/Ding0MHD021}
X.~Ding, X.~Zhang, N.~Ma, J.~Han, G.~Ding, and J.~Sun, ``Rep{VGG}: Making
  vgg-style convnets great again,'' in \emph{Conference on Computer Vision and
  Pattern Recognition, {CVPR}}.\hskip 1em plus 0.5em minus 0.4em\relax Computer
  Vision Foundation / {IEEE}, 2021, pp. 13\,733--13\,742.

\bibitem{DBLP:conf/cvpr/Ding0HD21}
X.~Ding, X.~Zhang, J.~Han, and G.~Ding, ``Diverse branch block: Building a
  convolution as an inception-like unit,'' in \emph{Conference on Computer
  Vision and Pattern Recognition, {CVPR}}.\hskip 1em plus 0.5em minus
  0.4em\relax Computer Vision Foundation / {IEEE}, 2021, pp. 10\,886--10\,895.

\bibitem{DBLP:conf/mm/ZhangZZ21}
X.~Zhang, H.~Zeng, and L.~Zhang, ``Edge-oriented convolution block for
  real-time super resolution on mobile devices,'' in \emph{{ACM} International
  Conference on Multimedia}.\hskip 1em plus 0.5em minus 0.4em\relax {ACM},
  2021, pp. 4034--4043.

\bibitem{DBLP:conf/mm/WangDS22}
X.~Wang, C.~Dong, and Y.~Shan, ``Rep{SR}: Training efficient vgg-style
  super-resolution networks with structural re-parameterization and batch
  normalization,'' in \emph{{ACM} International Conference on
  Multimedia}.\hskip 1em plus 0.5em minus 0.4em\relax {ACM}, 2022, pp.
  2556--2564.

\bibitem{yuan2012automatic}
L.~Yuan and J.~Sun, ``Automatic exposure correction of consumer photographs,''
  in \emph{Computer Vision - {ECCV} - 12th European Conference on Computer
  Vision}, ser. Lecture Notes in Computer Science, vol. 7575.\hskip 1em plus
  0.5em minus 0.4em\relax Springer, 2012, pp. 771--785.

\bibitem{DBLP:conf/bmvc/LvLWL18}
F.~Lv, F.~Lu, J.~Wu, and C.~Lim, ``{MBLLEN:} low-light image/video enhancement
  using {CNN}s,'' in \emph{British Machine Vision Conference {BMVC}}.\hskip 1em
  plus 0.5em minus 0.4em\relax {BMVA} Press, 2018, p. 220.

\bibitem{DBLP:conf/bmvc/CuiL0SG00H22}
Z.~Cui, K.~Li, L.~Gu, S.~Su, P.~Gao, Z.~Jiang, Y.~Qiao, and T.~Harada, ``You
  only need 90k parameters to adapt light: a light weight transformer for image
  enhancement and exposure correction,'' in \emph{British Machine Vision
  Conference {BMVC}}.\hskip 1em plus 0.5em minus 0.4em\relax {BMVA} Press,
  2022, p. 238.

\bibitem{vonikakis2021busting}
\BIBentryALTinterwordspacing
V.~Vonikakis, ``Busting image enhancement and tonemapping algorithms,'' 2021.
  [Online]. Available: \url{https://sites.google.com/site/vonikakis/datasets}
\BIBentrySTDinterwordspacing

\bibitem{DBLP:journals/tip/MaZW15}
K.~Ma, K.~Zeng, and Z.~Wang, ``Perceptual quality assessment for multi-exposure
  image fusion,'' \emph{{IEEE} Trans. Image Process.}, vol.~24, no.~11, pp.
  3345--3356, 2015.

\bibitem{DBLP:conf/icip/LeeLK12}
C.~Lee, C.~Lee, and C.~Kim, ``Contrast enhancement based on layered difference
  representation,'' in \emph{19th {IEEE} International Conference on Image
  Processing, {ICIP}}.\hskip 1em plus 0.5em minus 0.4em\relax {IEEE}, 2012, pp.
  965--968.

\bibitem{Adobe}
\BIBentryALTinterwordspacing
``Color and camera raw.'' 2020. [Online]. Available: \url{https://helpx.
  adobe.com/ca/photoshop- elements/using/ color-camera-raw.html}
\BIBentrySTDinterwordspacing

\bibitem{kingma2014adam}
D.~P. Kingma and J.~Ba, ``Adam: {A} method for stochastic optimization,'' in
  \emph{3rd International Conference on Learning Representations, {ICLR}},
  2015.

\bibitem{DBLP:conf/iclr/LoshchilovH17}
I.~Loshchilov and F.~Hutter, ``{SGDR:} stochastic gradient descent with warm
  restarts,'' in \emph{5th International Conference on Learning
  Representations, {ICLR}}.\hskip 1em plus 0.5em minus 0.4em\relax
  OpenReview.net, 2017.

\bibitem{DBLP:journals/tip/CaiGZ18}
J.~Cai, S.~Gu, and L.~Zhang, ``Learning a deep single image contrast enhancer
  from multi-exposure images,'' \emph{{IEEE} Trans. Image Process.}, vol.~27,
  no.~4, pp. 2049--2062, 2018.

\end{thebibliography}

\end{document}